\documentclass[manuscript,screen]{acmart}
\AtBeginDocument{%
  \providecommand\BibTeX{{%
    Bib\TeX}}}

\setcopyright{acmlicensed}
\copyrightyear{2018}
\acmYear{2018}
\acmDOI{XXXXXXX.XXXXXXX}
\acmConference[Conference acronym 'XX]{}{June 03--05,
  2018}{Woodstock, NY}
\acmISBN{978-1-4503-XXXX-X/2018/06}

\usepackage{booktabs}
\usepackage{multirow}
\usepackage{threeparttable}
\usepackage{color, colortbl}
\usepackage[commandnameprefix=always]{changes}
\usepackage{subcaption}
\usepackage{wrapfig}
\usepackage{diagbox}
\usepackage{arydshln} 
\usepackage{makecell}
\usepackage{enumitem}
\usepackage[graphicx]{realboxes}
\usepackage{balance}
\usepackage{textcomp}
\usepackage[table]{xcolor}
\usepackage{url}
\usepackage{booktabs}
\usepackage{array}
\usepackage{caption}
\usepackage[T1]{fontenc}
\usepackage{forest}
\usepackage{longtable}
\usepackage{wrapfig}
\usepackage{mathtools}

\PassOptionsToPackage{colorlinks=true, linkcolor=blue, citecolor=blue, urlcolor=blue}{hyperref}

\def\BibTeX{{\rm B\kern-.05em{\sc i\kern-.025em b}\kern-.08em
    T\kern-.1667em\lower.7ex\hbox{E}\kern-.125emX}}

\usepackage{pifont}

\begin{document}

\title{Harnessing AI for Inverse Partial Differential Equation Problems: Past, Present, and Prospects}

\author{Zhentao Tan}
\email{tanzhentao@zju.edu.cn}
\affiliation{%
  \institution{Collaborative Innovation Center of Artificial Intelligence
(CCAI), Zhejiang University}
  \city{Hangzhou}
  \state{Zhejiang}
  \country{China}
}
\author{Yuze Hao}
\email{yuzehao@zju.edu.cn}
\affiliation{%
  \institution{Collaborative Innovation Center of Artificial Intelligence
(CCAI), Zhejiang University}
  \city{Hangzhou}
  \state{Zhejiang}
  \country{China}
}

\author{Boyi Zou}
\email{boyizou@zju.edu.cn}
\affiliation{%
  \institution{School of Mathematical Sciences, Zhejiang University}
  \city{Hangzhou}
  \state{Zhejiang}
  \country{China}
}

\author{Mingsheng Long}
\email{mingsheng@tsinghua.edu.cn}
\affiliation{%
  \institution{Tsinghua University}
  \city{Beijing}
  \country{China}
}

\author{Yi Yang}
\email{yangyics@zju.edu.cn}
\affiliation{%
  \institution{Collaborative Innovation Center of Artificial Intelligence
(CCAI), Zhejiang University}
  \city{Hangzhou}
  \state{Zhejiang}
  \country{China}
}

\author{Gang Bao}
\email{baog@zju.edu.cn}
\affiliation{%
  \institution{Center for Interdisciplinary Applied Mathematics, School of Mathematical Sciences, Zhejiang University}
  \city{Hangzhou}
  \state{Zhejiang}
  \country{China}
}

\begin{abstract}
  Solving inverse partial differential equation (PDE) problems is a fundamental topic in scientific research due to its broad significance across a wide range of real-world applications. Inverse PDE problems arise across medical imaging, geophysics, materials science, and aerodynamics, where the goal is to infer hidden causes, design structures, or control physical states. In this paper, we provide a comprehensive review of recent advances in solving inverse PDE problems using artificial intelligence (AI). We first introduce the basic formulation, key challenges, and traditional numerical foundations of inverse PDE problems, and then organize it into three major categories: inverse problems, inverse design, and control problems. For each category, we further present a methodological paradigms, and review representative state-of-the-art approaches from recent years. We then summarize representative applications across scientific and industrial domains, including mechanical systems, aerodynamic problems, thermal systems, full-waveform inversion, system identification, and medical imaging. Finally, we discuss open challenges and future prospects, such as physics-informed architectures, limited real-world data, uncertainty quantification, and inverse foundation models. This survey aims to provide the first unified and systematic perspective on AI for inverse PDE problems, demonstrating how modern learning-based methods are reshaping inverse problems, inverse design, and control problems in PDE-governed systems.
\end{abstract}

\ccsdesc[500]{General and reference~Surveys and overviews}
\ccsdesc[500]{Computing methodologies~Artificial intelligence}
\ccsdesc[500]{Computing methodologies~Modeling and simulation}
\ccsdesc[500]{Mathematics of computing~Partial differential equations}
\ccsdesc[500]{Applied computing~Physical sciences and engineering}

\keywords{Inverse Problems, Inverse Design, Control Problems}


\maketitle

\section{Introduction}
Partial differential equations (PDEs) are one of the central mathematical tools for modeling physical, biological, and engineered systems. They describe how spatially and temporally varying fields evolve under conservation laws, constitutive relations, and external constraints. As a result, PDEs play a foundational role across a broad range of disciplines, including fluid dynamics~\cite{balla2022inverse, anand2024novel, elrefaie2024drivaernet, elrefaie2024drivaernet++, glaws2022invertible, dussauge2023reinforcement}, geophysics~\cite{zhang2020data, tang2021deep, kang2026implicit, zhang2025lateral, zhang2026enhancing}, electromagnetics~\cite{denker2025deep, de2025extension, chen2026data, guo2025warm, cao2025diff}, mechanical systems~\cite{haghighat2021physics, senhora2022machine, deng2022self, yang2024guided, bastek2023inverse}.

In modern scientific and industrial practice, PDE-based models are used not only for forward simulation, but also as a computational foundation for parameter estimation, system design, and decision-making, all of which can be collectively formulated as inverse PDE problems. At the same time, the growing availability of sensing data~\cite{santos2023development, jean2021copernicus, jasak2009openfoam}, advances in high-performance computing, and the rapid development of machine learning have greatly expanded both the scale and the scope of PDE-driven modeling, making inverse problems for PDE systems an increasingly important research direction.

In contrast to forward PDE problems, where the governing equations, parameters, and initial and boundary conditions are assumed to be known, and the goal is to solve for the system state, inverse PDE problems aim to recover unknown quantities from incomplete observations or prescribed targets. These unknowns may correspond to physical coefficients, source terms, initial or boundary conditions, latent states, control variables, and structural designs. Such problems are of fundamental importance due to their broad relevance in real-world applications: in medical imaging, one seeks to reconstruct hidden tissues from measurements; in geophysics, one infers subsurface properties from seismic responses; in engineering, one searches for structures that satisfy desired physical behaviors; and in control problems, one seeks actions that steer a PDE-governed system toward a target state. However, inverse PDE problems are also substantially more challenging than their forward counterparts. They are often ill-posed, with solutions that may be non-unique, unstable, or highly sensitive to sparse and noisy observations. In addition, the forward operator is frequently nonlinear, high-dimensional, and computationally expensive to evaluate repeatedly, making classical iterative solvers difficult to scale.

\begin{figure*}[!t]
\centering
\includegraphics[width=0.95\textwidth]{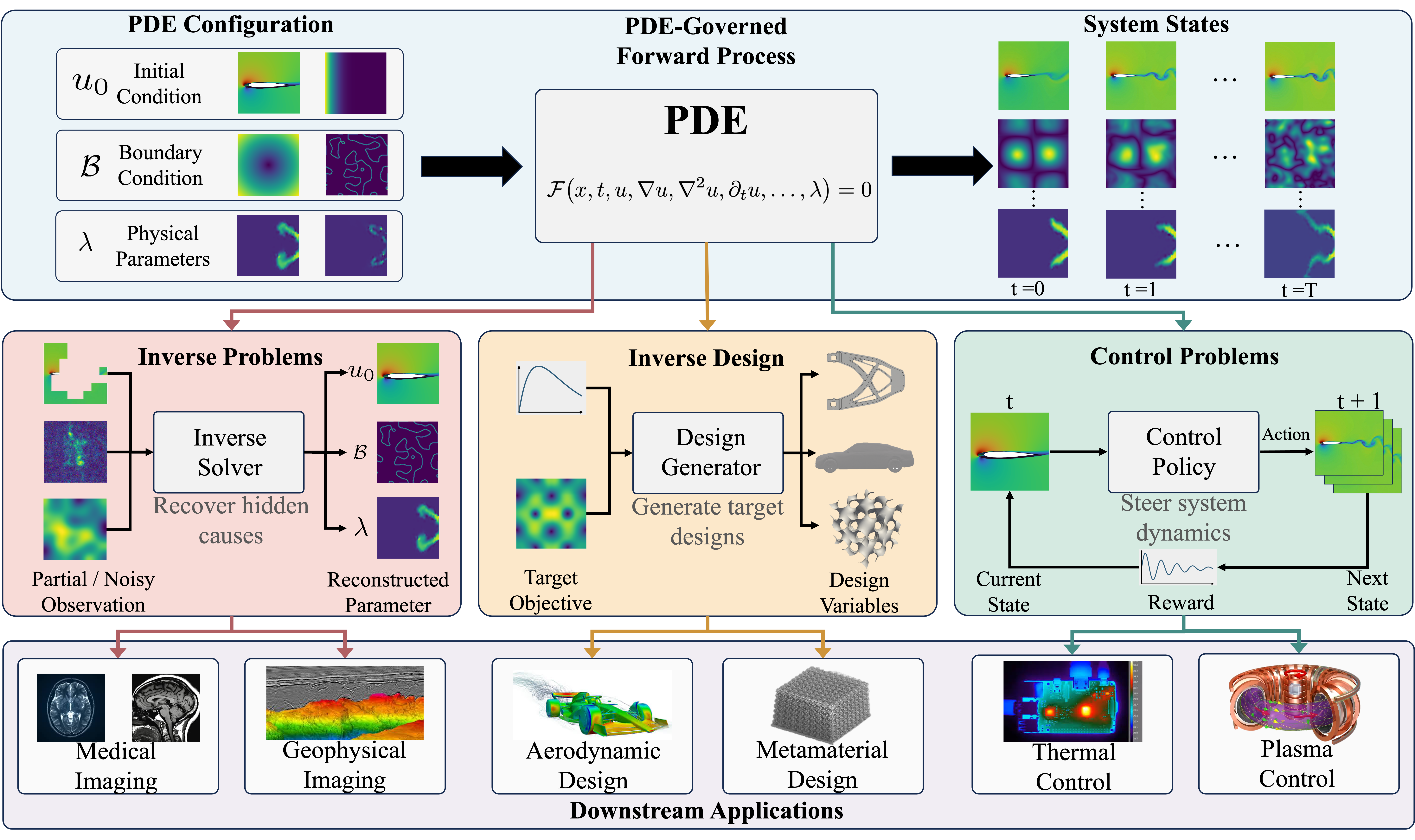}
\caption{Overview of AI-based inverse PDE problems. 
\textbf{The upper row} illustrates the forward PDE process, where system configurations, including initial conditions, boundary conditions, and physical parameters, evolve into system states. 
\textbf{The middle row} summarizes three major inverse tasks: inverse problems, inverse design, and control problems.
\textbf{The bottom row} shows representative downstream applications across various scenarios.}
\label{fig:Overview}
\end{figure*}

With the rapid advancement of AI in fields such as computer vision and natural language processing, its application to inverse PDE problems has emerged as a promising research direction. We decompose AI methods for inverse PDE problems into three major problem classes: inverse problems, inverse design, and control problems. Although these settings differ in formulation and downstream objectives, they share a common principle: the goal is not merely to simulate a PDE system, but to infer, optimize, or decide over hidden variables under PDE constraints. Inverse problems primarily concern the recovery of unknown physical quantities from partial observations. Inverse design, by contrast, focuses on identifying structures, parameters, or configurations that yield desired behaviors. Control problems involve decision-making over time, with the objective of determining control policies or action sequences for PDE-governed dynamical systems. This three-part taxonomy provides a unified way to compare methods that otherwise appear in separate communities, and it highlights how recent AI advances have reshaped inverse reasoning from direct regression to optimization, operator learning, generative modeling, and sequential decision-making.

\definecolor{steelblue}{HTML}{5B8FA8}
\definecolor{mintgreen}{HTML}{9DC4B0}
\definecolor{amberorange}{HTML}{F5A84B}
\definecolor{dustyred}{HTML}{B85450}
\definecolor{lavender}{HTML}{9B7BB0}

\colorlet{steelblue-light}{steelblue!25}
\colorlet{mintgreen-light}{mintgreen!30}
\colorlet{amberorange-light}{amberorange!25}
\colorlet{dustyred-light}{dustyred!20}
\colorlet{lavender-light}{lavender!25}

\usetikzlibrary{arrows.meta,positioning,calc,patterns,shadows.blur}
\usetikzlibrary{decorations.pathmorphing}

\definecolor{rootblue}{HTML}{5A8DEE}

\definecolor{underC}{HTML}{C9A227}
\definecolor{reasonC}{HTML}{8AA05A}
\definecolor{genC}{HTML}{63B38A}
\definecolor{appC}{HTML}{4DB6AC}

\colorlet{underLight}{underC!18}
\colorlet{reasonLight}{reasonC!18}
\colorlet{genLight}{genC!18}
\colorlet{appLight}{appC!18}

\definecolor{hidden-draw}{RGB}{120,120,120}

\tikzstyle{my-box}=[
    rectangle,
    draw=hidden-draw,
    rounded corners,
    align=left,
    text opacity=1,
    minimum height=1.5em,
    minimum width=5em,
    inner sep=2pt,
    fill opacity=.8,
    line width=0.8pt,
]
\tikzstyle{leaf-head}=[my-box, minimum height=1.5em,
    draw=steelblue, fill=steelblue-light,
    text=black, font=\normalsize,
    inner xsep=2pt, inner ysep=4pt, line width=0.8pt,
]
\tikzstyle{leaf-task}=[my-box, minimum height=2.5em,
    draw=dustyred, fill=dustyred-light,
    text=black, font=\normalsize,
    inner xsep=2pt, inner ysep=4pt, line width=0.8pt,
]
\tikzstyle{modelnode-task}=[my-box, minimum height=1.5em,
    draw=dustyred, fill=dustyred-light,
    text=black, font=\normalsize,
    inner xsep=2pt, inner ysep=4pt, line width=0.8pt,
]
\tikzstyle{leaf-paradigms}=[my-box, minimum height=2.5em,
    draw=amberorange, fill=amberorange-light,
    text=black, font=\normalsize,
    inner xsep=2pt, inner ysep=4pt, line width=0.8pt,
]
\tikzstyle{modelnode-paradigms}=[my-box, minimum height=1.5em,
    draw=amberorange, fill=amberorange-light,
    text=black, font=\normalsize,
    inner xsep=2pt, inner ysep=4pt, line width=0.8pt,
]
\tikzstyle{leaf-others}=[my-box, minimum height=2.5em,
    draw=mintgreen, fill=mintgreen-light,
    text=black, font=\normalsize,
    inner xsep=2pt, inner ysep=4pt, line width=0.8pt,
]
\tikzstyle{modelnode-others}=[my-box, minimum height=1.5em,
    draw=mintgreen, fill=mintgreen-light,
    text=black, font=\normalsize,
    inner xsep=2pt, inner ysep=4pt, line width=0.8pt,
]
\tikzstyle{leaf-application}=[my-box, minimum height=2.5em,
    draw=lavender, fill=lavender-light,
    text=black, font=\normalsize,
    inner xsep=2pt, inner ysep=4pt, line width=0.8pt,
]
\tikzstyle{modelnode-application}=[my-box, minimum height=1.5em,
    draw=lavender, fill=lavender-light,
    text=black, font=\normalsize,
    inner xsep=2pt, inner ysep=4pt, line width=0.8pt,
]

\begin{figure*}[!th]
    \centering
    \resizebox{1\textwidth}{!}
    {
        \begin{forest}
            for tree={
                grow=east,
                reversed=true,
                anchor=base west,
                parent anchor=east,
                child anchor=west,
                base=left,
                font=\normalsize,
                rectangle,
                draw=hidden-draw,
                rounded corners,
                align=left,
                minimum width=1em,
                edge+={darkgray, line width=1pt},
                s sep=1pt,
                inner xsep=0pt,
                inner ysep=2pt,
                line width=0.8pt,
                ver/.style={rotate=90, child anchor=north, parent anchor=south, anchor=center},
            },
            [
                AI for Inverse PDE Problems, leaf-head, ver
                [
                    \S \ref{sec:IP} Inverse Problems, leaf-task, ver [ 
                    \S \ref{dd ip} Data-Driven, leaf-task, text width=9em
                    [
                        \S \ref{direct mapping ip} Inverse-\\Mapping, leaf-task, text width=8em
                        [\textbf{INN}~\cite{ardizzoneanalyzing}{, }\textbf{c-INN}~\cite{kruse2021benchmarking}{, }\textbf{VINA}~\cite{shekhar2022vina}{, }\textbf{i-ResNets}~\cite{behrmann2019invertible}{, }\textbf{i-AE}~\cite{teng2019invertible}{, }\textbf{MDN}~\cite{kruse2020technical}{, }\textbf{LNO}~\cite{wang2024latent}{, }\\ \textbf{Liu \textit{et al.}}~\cite{liu2024bathymetry}{, }\textbf{NIO}~\cite{molinaro2023neural}{, }\textbf{iFNO}~\cite{longinvertible}{, }\textbf{Zhao \textit{et al.}~\cite{zhao2022learning}}
                        {, }\textbf{IGNO}~\cite{zang2025unified}{, }\textbf{Jiao \textit{et al.}}~\cite{jiao2026solving}, modelnode-task, text width=38em]
                    ]
                    [
                        \S \ref{pde constrained ip} PDE-\\Constrained, leaf-task, text width=8em
                        [\textbf{PINN}~\cite{raissi2019physics}{, }\textbf{gPINN}~\cite{yu2022gradient}{, }\textbf{WANs}~\cite{bao2020numerical}{, }\textbf{DG-PINN}~\cite{zhou2024data}{, }\textbf{PIED}~\cite{hemachandrapied}{, }\textbf{cPINN}~\cite{chen2020physics}{, }\textbf{PI}-\\ \textbf{KAN}~\cite{perez2025physics}, modelnode-task, text width=38em]
                    ]
                    [
                        \S \ref{surrogate ip} Surrogate-\\Based, leaf-task, text width=8em
                        [\textbf{Neural-Adjoint}~\cite{ren2020benchmarking}{, }\textbf{PETAL}~\cite{jin2023petal}{, }\textbf{Weidner \textit{et al.}}~\cite{weidner2024learnable}{, }, modelnode-task, text width=38em]
                    ]
                    ]
                    [
                        \S \ref{TJ ip} Inverse dynam-\\ics, leaf-task, text width=9em
                        [\textbf{NeuroFluid}~\cite{guan2022neurofluid}{, }\textbf{HyFluid}~\cite{xu2025hybrid}{, }\textbf{GIC}~\cite{cai2024gic}{, }\textbf{MASIV}~\cite{zhao2025toward}{, }\textbf{SciML}~\cite{liudata}
                        {, }\textbf{ProJo4D}~\cite{rho2025projo4d}{, }\textbf{Deng \textit{et al.}}~\cite{denglearning}{, }\\ \textbf{IFD}~\cite{liu2023inferring}{, }\textbf{Zhao \textit{et al.}}~\cite{zhao20253d}{, }\textbf{Zhu \textit{et al.}}~\cite{zhulatent}, modelnode-task, text width=47.7em]
                    ]
                    [
                        \S \ref{GB ip} Generative-\\Based, leaf-task, text width=9em
                        [
                            Diffusion-based, leaf-task, text width=9em
                            [\textbf{DPS}~\cite{chung2022diffusion}{, }\textbf{DiffusionPDE}~\cite{huang2024diffusionpde}{, }\textbf{FunDPS}~\cite{yao2025guided}{, }\textbf{DDIS}~\cite{lin2026decoupled}{, }\textbf{Fun-DDPS}~\cite{ju2026function}, modelnode-task, text width=37em]
                        ]
                        [
                            GAN-based, leaf-task, text width=9em
                            [\textbf{GANO}~\cite{rahman2022generative}{, }\textbf{GAROM}~\cite{coscia2024generative}, modelnode-task, text width=37em]
                        ]
                    ]
                ]
                [
                    \S \ref{sec:ID} Inverse Design, leaf-paradigms, ver
                    [
                        \S \ref{ID:Opt} Optimization-\\Based, leaf-paradigms, text width=9em
                        [
                            \S \ref{ID:Opt:PhysicsLoop} Physics-in-the-loop, leaf-paradigms, text width=16em
                            [\textbf{JAX-Fluids}~\cite{bezgin2023jax}{, }\textbf{NeuralFluid}~\cite{li2024neuralfluid}{, }\textbf{$\phi$Flow}~\cite{holl2024bf}{, }\textbf{Luce \textit{et al.}}~\cite{luce2024merging}, modelnode-paradigms, text width=30em]
                        ]
                        [
                            \S \ref{ID:Opt:surrogate} Learned-Model\\-Assisted, leaf-paradigms, text width=16em
                            [\textbf{Allen \textit{et al.}}~\cite{allen2022inverse}{, }\textbf{Lu \textit{et al.}}~\cite{lu2022multifidelity}{, }\textbf{Augenstein \textit{et al.}}~\cite{augenstein2023neural}, modelnode-paradigms, text width=30em]
                        ]
                    ]
                    [
                        \S \ref{ID:Dir} Direct Inverse\\Design, leaf-paradigms, text width=9em
                        [\textbf{Liu \textit{et al.}}~\cite{liu2018training}{, }\textbf{Tahersima \textit{et al.}}~\cite{tahersima2019deep}{, }\textbf{Unni \textit{et al.}}~\cite{unni2021mixture}{, }\textbf{Sekar \textit{et al.}}~\cite{sekar2019inverse}{, }\textbf{Grbcic \textit{et al.}}~\cite{grbvcic2025inverse}, modelnode-paradigms, text width=47.6em]
                    ]
                    [
                        \S \ref{ID:Gen} Generative\\Inverse Design, leaf-paradigms, text width=9em
                        [
                            \S \ref{ID:Gen:GAN} Latent-Variable\&Adversarial, leaf-paradigms, text width=16em
                            [\textbf{GLOnet}~\cite{jiang2020simulator}{, }\textbf{PcDGAN} ~\cite{heyrani2021pcdgan}{, }\textbf{Wang \textit{et al.}}~\cite{wang2020deep}{, }\textbf{A-CVAE}~\cite{tang2020generative}, modelnode-paradigms, text width=30em]
                        ]
                        [
                            \S \ref{ID:Gen:Diffusion} Diffusion-Based, leaf-paradigms, text width=16em
                            [\textbf{ZeoDiff}~\cite{park2024inverse}{, }\textbf{TopoDiff}~\cite{maze2023diffusion}{, }\textbf{Cindm }~\cite{wu2024compositional}{, }
                            \textbf{MetaGen}~\cite{hen2025inverse}, modelnode-paradigms, text width=30em]
                        ]
                    ]
                ]
                [
                    \S \ref{sec:IC} Control Problems, leaf-others, ver
                    [
                        \S \ref{IC:Opt} Optimization-\\Based, leaf-others, text width=9em
                        [
                            \S \ref{IC:Opt:PDECons} Physics-in-the-loop Control, leaf-others, text width=16em
                            [\textbf{CFE}~\cite{holl2020learning}{, }\textbf{Control PINNs}~\cite{barry2025physics}{, }\textbf{PhysNet}~\cite{gokhale2022physics}{, }\textbf{Mowlavi \textit{et al.}}~\cite{mowlavi2023optimal}, modelnode-others, text width=30em]
                        ]
                        [
                            \S \ref{IC:Opt:Surrogate} Learned-Model-Assisted, leaf-others, text width=16em
                            [\textbf{DMDc}~\cite{proctor2016dynamic}{, }\textbf{K-ROMs}~\cite{peitz2019koopman}{, }\textbf{MS-DeepONet}~\cite{de2025deep}{, }\textbf{Korda \textit{et al.}}~\cite{korda2018linear}, modelnode-others, text width=30em]
                        ]
                    ]
                    [
                        \S \ref{IC:Amor} Offline Trained\\Control, leaf-others, text width=9em
                        [
                            \S \ref{IC:Amor:Operator} Operator Learning, leaf-others, text width=16em
                            [\textbf{OCPs}~\cite{feng2025optimal}{, }\textbf{Wang \textit{et al.}}~\cite{wang2021fast}{, }\textbf{Bhan \textit{et al.}}~\cite{bhan2023neural}{, }\textbf{Krstic \textit{et al.}}~\cite{krstic2024neural}, modelnode-others, text width=30em]
                        ]
                        [
                            \S \ref{IC:Amor:Imitation} Imitation Learning, leaf-others, text width=16em
                            [\textbf{Neural-HJB}~\cite{verma2025neural}{, }\textbf{BEAR}~\cite{mulayim2025impact}, modelnode-others, text width=30em]
                        ]
                        [
                            \S \ref{IC:Amor:Generative} Generative Policy Learning, leaf-others, text width=16em
                            [\textbf{DiffPhyCon}~\cite{wei2024diffphycon}{, }\textbf{CL-DiffPhyCon}~\cite{wei2024cl}{, }\textbf{SafeDiffCon}~\cite{hu2025uncertain}, modelnode-others, text width=30em]
                        ]
                    ]
                    [
                        \S \ref{IC:RL} Reinforcement\\Learning, leaf-others, text width=9em
                        [\textbf{MARL-DRL}~\cite{guastoni2023deep}{, }\textbf{SmartSOD2D}~\cite{font2025deep}{, }\textbf{Invariant MARL}~\cite{vignon2023effective}{, }\textbf{Rabault \textit{et al.}}~\cite{rabault2019artificial}{, }\textbf{Degrave \textit{et al.}}~\cite{degrave2022magnetic}, modelnode-others, text width=47.6em]
                    ]
                ]
                [
                    \S \ref{sec:APP} Applications, leaf-application, ver
                    [
                        \S \ref{APP:Mechanical} Mechanical Systems, leaf-application, text width=14em
                        [\textbf{Haghighat \textit{et al.}}~\cite{haghighat2021physics}{, }\textbf{Senhora \textit{et al.}}~\cite{senhora2022machine}{, }\textbf{Deng \textit{et al.}}~\cite{deng2022self}{, }\textbf{Yang \textit{et al.}}~\cite{yang2024guided}, modelnode-application, text width=42.5em]
                    ]
                    [
                        \S \ref{APP:Aerodynamic} Aerodynamic Problems, leaf-application, text width=14em
                        [\textbf{Balla \textit{et al.}}~\cite{balla2022inverse}{, }\textbf{Anand \textit{et al.}}~\cite{anand2024novel}{, }\textbf{Bezgin \textit{et al.}}~\cite{bezgin2023jax}{, }\textbf{Wang \textit{et al.}}~\cite{wang2023deep}{, }\textbf{Elrefaie \textit{et al.}}~\cite{elrefaie2024drivaernet}{, }\\ \textbf{Glaws \textit{et al.}}~\cite{glaws2022invertible}{, }\textbf{Dussauge \textit{et al.}}~\cite{dussauge2023reinforcement}, modelnode-application, text width=42.5em]
                    ]
                    [
                        \S \ref{APP:Thermal} Thermal Systems, leaf-application, text width=14em
                        [\textbf{Cai \textit{et al.}}~\cite{cai2021physics}{, } \textbf{Billah \textit{et al.}}~\cite{billah2023physics}{, }\textbf{Xu \textit{et al.}}~\cite{xu2023physics}{, }\textbf{Gokhale \textit{et al.}}~\cite{gokhale2022physics}{, }\textbf{Zhang \textit{et al.}}~\cite{zhang2024physics}{, }\\ \textbf{Majumdar \textit{et al.}}~\cite{majumdar2025hxpinn}, modelnode-application, text width=42.5em]
                    ]
                    [
                        \S \ref{APP:Full-waveform} Full-Waveform Inversion, leaf-application, text width=14em
                        [\textbf{Zhang \textit{et al.}}~\cite{zhang2020data}{, } \textbf{Tang \textit{et al.}}~\cite{tang2021deep}{, }\textbf{Wu \textit{et al.}}~\cite{wu2019inversionnet}{, }\textbf{Li \textit{et al.}}~\cite{li2025towards}{, }\textbf{Kang \textit{et al.}}~\cite{kang2026implicit}, modelnode-application, text width=42.5em]
                    ]
                    [
                        \S \ref{APP:Dynamical} System Identification, leaf-application, text width=14em
                        [\textbf{PAC-NeRF}~\cite{li2023pac}{, }\textbf{Lu \textit{et al.}}~\cite{lu2022discovering}{, }\textbf{Yang \textit{et al.}}~\cite{yang2025differentiable}, modelnode-application, text width=42.5em]
                    ]
                    [
                        \S \ref{APP:Medical} Medical Imaging, leaf-application, text width=14em
                        [\textbf{Song \textit{et al.}}~\cite{songsolving}{, }\textbf{DPS}~\cite{chung2022diffusion}{, }\textbf{DDS}~\cite{chung2023decomposed}{, }\textbf{$\Pi$GDM}~\cite{song2023pseudoinverse}{, }\textbf{DOLCE}~\cite{liu2023dolce}, modelnode-application, text width=42.5em]
                    ]
                ]
            ]
        \end{forest}
    }
    \caption{Taxonomy of AI-based methods for inverse PDE problems, organized along three main axes: inverse problems, inverse design, and control problems, with representative applications reviewed separately.}
    \label{fig_tax}
\vspace{-0.3cm}
\end{figure*}
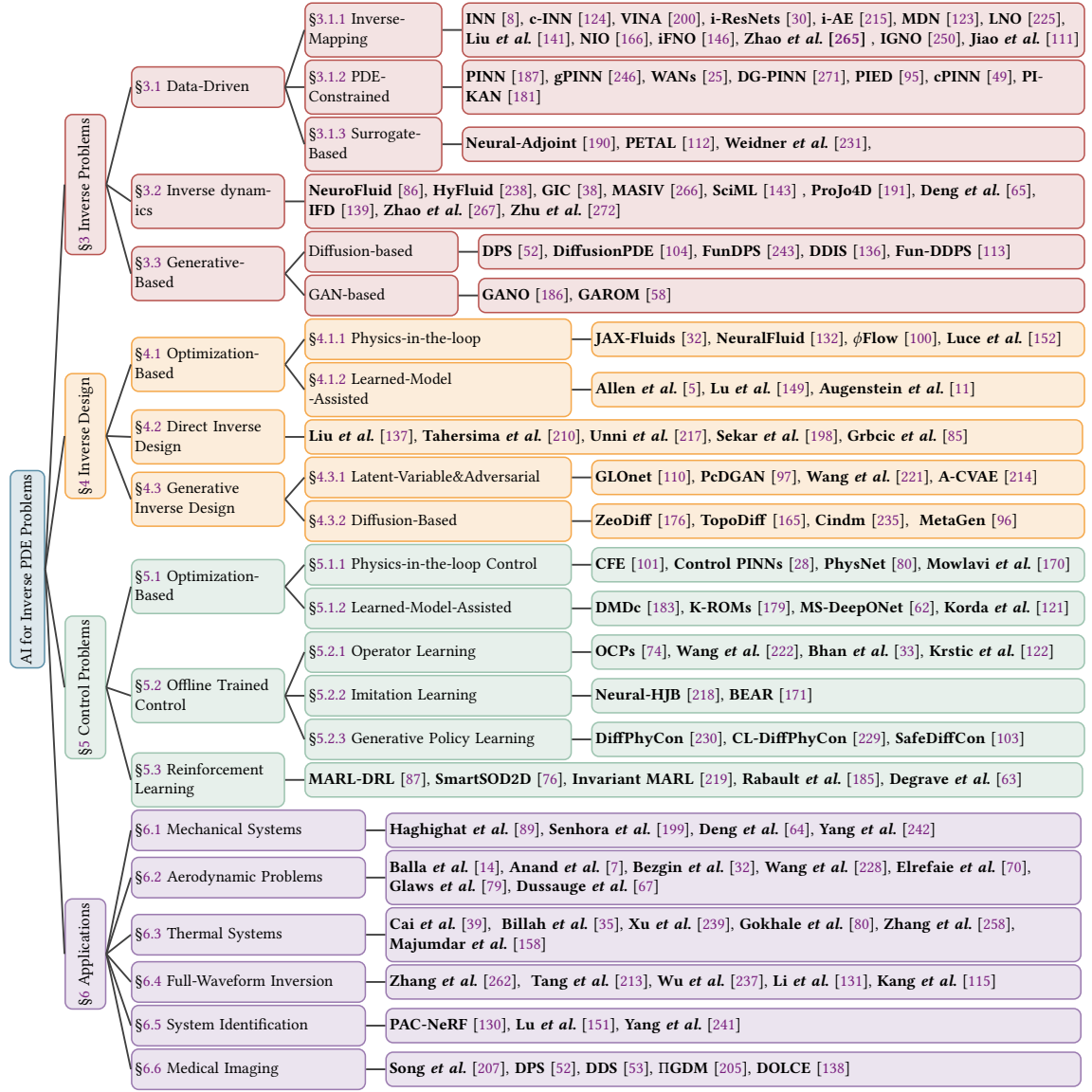


The remainder of this paper is organized as follows. Section~\ref{sec:Preliminary} presents the basic formulations of inverse problems for PDEs, and briefly reviews classical numerical methods for the associated inverse problems. Section~\ref{sec:IP} surveys AI methods for inverse problems, beginning with general data-driven paradigms, including \emph{inverse-mapping}, \emph{PDE-constrained}, and \emph{surrogate-based} methods. We further cover \emph{inverse dynamics} methods and recent \emph{generative-based} methods. Section~\ref{sec:ID} and Section~\ref{sec:IC} review AI methods for inverse design and control problems, respectively. Beyond methodology, we also discuss task-specific considerations in these two settings. For inverse design, we emphasize \emph{design space representation}, which is central to design tasks. For control problems, we highlight the \emph{temporal and feedback structure of PDE control problems}, which is essential to the development of effective control strategies. Finally, Section~\ref{sec:APP} presents representative scientific and industrial applications to illustrate how these AI-driven methods for inverse PDEs are deployed in practice, and we conclude by outlining several open challenges and future prospects. An overview of inverse PDE problems is shown in Fig.~\ref{fig:Overview}. Notably, several recent surveys~\cite{wang2024recent,zhang2025artificial} mainly focus on forward simulation for PDEs. In contrast, our survey provides the first comprehensive review dedicated to inverse PDE problems.

\section{Preliminary}\label{sec:Preliminary}

\subsection{Partial Differential Equations}

Partial differential equations (PDEs) describe how an unknown field varies with respect to multiple independent variables, such as space and time. They provide a fundamental framework for modeling a wide range of physical phenomena, including fluid flow~\cite{li2024neuralfluid}, wave propagation~\cite{kang2026implicit}, electromagnetics~\cite{denker2025deep}, and elasticity~\cite{bao2018inverse}. In general, a PDE can be written as
\begin{equation}
\label{general pde}
\mathcal{F}\big(x,t,u,\nabla u,\nabla^2 u,\partial_t u,\ldots, \lambda_p\big)=0,
\end{equation}
where $x \in \Omega \subset \mathbb{R}^n$ denotes the spatial coordinate, $t \in [0,T]$ denotes time, and $u=u(x,t)$ denotes the solution field, and \(\lambda_p\in\Lambda_p\) denotes the physical parameters, with \(\Lambda_p\) being the corresponding parameter space. The operator $\mathcal{F}$ is a differential operator that may involve spatial derivatives, temporal derivatives, and problem-dependent coefficients or source terms.

A typical PDE system is accompanied by initial and boundary conditions to ensure well-posedness. These constraints are commonly written as
\begin{equation}
\label{boundary}
\mathcal{B}[u](x,t)=0, \quad (x,t)\in \partial\Omega \times [0,T],
\end{equation}
\begin{equation}
\label{initial}
u(x,0)=u_0(x), \quad x \in \Omega,
\end{equation}
where $\mathcal{B}$ denotes the boundary operator and $u_0$ is the prescribed initial condition. 
In scientific computing, solving a PDE typically means estimating the unknown field $u$ that satisfies both the governing equation and the associated initial or boundary constraints. 

\subsection{Inverse Partial Differential Equation Problems}

For inverse PDE problems, we broadly categorize the problems into \textbf{inverse problems}, \textbf{inverse design}, and \textbf{control problems}. Although these settings differ in objectives and formulations, they share a common motivation: to reason backward through PDE-governed systems by inferring hidden causes, synthesizing design variables, or determining control actions under physical constraints.

Several canonical PDE systems are frequently adopted as benchmark problems in this area, including Darcy flow, the Poisson equation, the Helmholtz equation, the Burgers equation, and the Navier--Stokes equations with various initial and boundary conditions~\cite{bao2021recovering,takamoto2022pdebench,lu2021deepxde}. These examples span a wide range of elliptic, parabolic, and hyperbolic PDEs and have served as standard benchmarks for the development and evaluation of inverse PDE solvers.

Beyond the commonly used benchmark equations mentioned above, a broader class of PDE systems also arises in inverse problems, inverse design, and control problems. These include heat equations~\cite{billah2023physics}, advection--diffusion and reaction--diffusion systems~\cite{takamoto2022pdebench}, wave and acoustic equations~\cite{bao2009stability}, elasticity equations~\cite{bao2018inverse}, shallow water equations~\cite{takamoto2022pdebench}, Maxwell's equations~\cite{bao2004inverse,bao2022maxwell}, conductivity equations~\cite{augenstein2023neural} in electrical impedance tomography, as well as biologically and chemically motivated systems such as Keller--Segel~\cite{hillen2009user}, monodomain/bidomain~\cite{franzone2014mathematical}, and radiative transfer equations~\cite{peraiah2002introduction}. Together, these PDEs cover a wide spectrum of elliptic, parabolic, and hyperbolic systems, which reflect the diversity of real-world applications in science, engineering, and medicine.

\subsection{Traditional Methods}
\label{subsec:tradition}

Traditional numerical methods for solving forward problems include finite difference methods, finite element methods, finite volume methods, and spectral methods. Before the emergence of modern data-driven approaches, inverse PDE problems were primarily studied within the framework of nonlinear ill-posed operator equations and PDE-constrained optimization. In this setting, the goal is to recover unknown physical parameters, initial conditions, or boundary conditions from PDE solutions or sensor observations \(u_{\mathrm{obs}}\). A standard formulation is
\begin{equation}
\label{inverse optimize}
\min_{\lambda}\;
\frac{1}{2}\|u_\lambda-u_{obs}||^2
+
\frac{\beta}{2}\mathcal{R}(\lambda),
\end{equation}
where $u_\lambda$ denotes the solution of equation~\eqref{general pde} for a given $\lambda_p$ under equations~\eqref{boundary} and \eqref{initial}; $\lambda = (\lambda_p,u_0,\mathcal{B})$ denotes the unknown parameters; $\mathcal{R}(\lambda)$ is a regularization functional; and $\beta>0$ is the regularization parameter. The central difficulty is that inverse PDE problems are typically ill-posed: solutions may fail to exist, lack uniqueness, or may be highly unstable under data perturbations. Classical numerical methods therefore rely on three closely related ideas: regularization to promote stability, linearization to make optimization tractable, and structure-exploiting numerical schemes tailored to specific classes of PDEs.

A classical line of work for inverse PDE problems is based on nonlinear least-squares methods. Levenberg \textit{et al.}~\cite{levenberg1944method} introduced a damped update scheme, and Marquardt \textit{et al.}~\cite{marquardt1963algorithm} later improved it by bridging Gauss--Newton and gradient descent. Though originally developed for general nonlinear optimization, these methods became standard tools for inverse PDE problems after discretization, especially when local linearization is effective. In parallel, regularization methods, particularly Tikhonov regularization~\cite{golub1999tikhonov}, were widely adopted for ill-posed inverse problems.

For distributed-parameter systems, these ideas were later developed into variational and PDE-constrained formulations. Chavent \textit{et al.}~\cite{chavent1979identification} formulated parameter identification as minimizing the mismatch between measured and simulated states, while also discussing implementation problems. Ito \textit{et al.}~\cite{ito1990augmented} then introduced an augmented Lagrangian method for parameter estimation in elliptic systems, combining output least-squares with equation-error formulations. Variational methods~\cite{acar1993identification} were also studied for coefficient identification in elliptic equations. Giordana \textit{et al.}~\cite{giordana1992numerical} studied more structured settings such as diffusion equations with piecewise constant coefficients. Traditional numerical methods for inverse PDE problems are usually formulated as regularized optimization problems, with gradients computed by sensitivity or adjoint analysis. However, simple regularization often fails to capture the intrinsic structure of the physical parameter space, and repeated calls to expensive forward solvers can make these methods inefficient in real-world applications.

To improve stability and reconstruction quality in inverse PDE problems, another representative line of traditional methods focuses on inverse scattering problems, where Helmholtz- or Maxwell-type equations are used to recover unknown media, interfaces, or scatterers from measured wave fields. Along this line of research, Bao \textit{et al.} developed a series of classical numerical methods for inverse medium scattering, stochastic scattering, and multi-frequency reconstruction~\cite{bao2005inverse,bao2010numerical,bao2013near,bao2015inverse}, typically relying on recursive linearization, frequency continuation, with notable improvements in reconstruction stability, accuracy, and computational effectiveness.

\begin{table*}[ht]
\centering
\scriptsize
\caption{Overview of AI methods for Inverse Problem.}
\label{tab:ip}
\vspace{-0.2cm}

\setlength{\tabcolsep}{4.5mm}{
\begin{tabular}{l|c|c|c}
\hline
\textbf{Methods} & \textbf{Source} & \textbf{Backbone} & \textbf{Tasks} \\ \hline
\multicolumn{4}{c}{\textbf{\ref{dd ip} Data-driven methods}} \\ \hline
\multicolumn{4}{l}{\textbf{\ref{direct mapping ip} Inverse-mapping}} \\ \hline
INN~\cite{ardizzoneanalyzing} & \textit{ICLR2019}  & MLP & Inverse kinematics  \\ \hline
iResNet~\cite{behrmann2019invertible} & \textit{ICML2019}  & CNN & Inverse image downsampling  \\ \hline
IAE~\cite{teng2019invertible} & \textit{Computation2019}  & CNN & Road video conversion  \\ \hline
cINN~\cite{kruse2021benchmarking} & \textit{Arxiv2021}  & MLP & Inverse kinematics, inverse Ballistics  \\ \hline
VINA~\cite{shekhar2022vina} & \textit{JMLR2022}  & CNN & Ocean-acoustic inversion  \\ \hline
MDN~\cite{kruse2020technical} & \textit{TechReport2020}  & CNN & Robot inverse kinematics  \\ \hline
Torfeh \textit{et al.}~\cite{torfeh2025probabilistic} & \textit{J. Phys. Photonics2025}  & CNN & Inverse fabrication-ready metasurfaces   \\ \hline
NIOs~\cite{molinaro2023neural} & \textit{ICML2023}  & MLP &  Inverse wave scattering, optical \& seismic imaging  \\ \hline
iFNO~\cite{longinvertible}  & \textit{ICLR2025}  & MLP & Wave, Navier-Stokes, Eikonal, Groundwater, Cell Growth  \\ \hline
LNO~\cite{wang2024latent}  & \textit{NeurIPS2025}  & MLP & Wave, Navier-Stokes, Eikonal, Groundwater, Cell Growth  \\ \hline
PI-DeepONet~\cite{jiao2026solving}  & \textit{SIAM2026}  & MLP & Wave, Navier-Stokes, Eikonal, Groundwater, Cell Growth  \\ \hline
IGNO~\cite{zang2025unified} & \textit{Arxiv2025}  & MLP & Electrical impedance tomography  \\ \hline
\multicolumn{4}{l}{\textbf{\ref{pde constrained ip} PDE-constrained}} \\ \hline
PINN~\cite{raissi2019physics} & \textit{J. Comput. Phys.2019}  & MLP & Allen-Cahn, Navier-Stokes, Schrodinger, Burgers  \\ \hline
cPINN~\cite{chen2020physics} & \textit{Opt. Express2020}  & MLP &  Helmholtz equation, inverse nanocylinder flow  \\ \hline
gPINN~\cite{yu2022gradient} & \textit{CMAME2022}  & MLP & Poisson, Diffusion–reaction, Burgers, Allen-Cahn \\ \hline
WANs~\cite{bao2020numerical} & \textit{Inverse Problems2020} & MLP & Inverse EIT, dynamic EIT \\ \hline
DG-PINNs~\cite{zhou2024data} & \textit{Arxiv2024}  & MLP & Wave equation, Navier–Stokes equation\\ \hline
WAF-PINN~\cite{zhang2025inverse} & \textit{Appl. Math. Model2025}  & MLP & Inverse (Poisson, Navier–Stokes, solid mechanics) \\ \hline
PIED~\cite{hemachandrapied} & \textit{ICLR2025}  & MLP & Wave, Navier-Stokes, Eikonal, Groundwater, Cell Growth  \\ \hline
PIKAN~\cite{perez2025physics} & \textit{Arxiv2025}  & KAN & Semi-infinite domain inverse problem \\ \hline
\multicolumn{4}{l}{\textbf{\ref{surrogate ip} Surrogate-based}} \\ \hline
NA~\cite{ren2020benchmarking} & \textit{NeurIPS2020}  & MLP & Ballistics, sine wave, robotic arm, meta material  \\ \hline
Zhao \textit{et al.}~\cite{zhao2022learning} & \textit{ICML2022}  & GNN &  Full wave inversion, Navier–Stokes, fluid assimilation \\ \hline
Liu \textit{et al.}~\cite{liu2024bathymetry} & \textit{WRR2024}  & CNN & Inverse shallow water equation \\ \hline
PETAL~\cite{jin2023petal} & \textit{NeurIPS2023}  & MLP & Inverse ocean acoustic tomography \\ \hline
Weidner \textit{et al.}~\cite{weidner2024learnable} & \textit{TMI2024}  & CNN & Inverse tumor growth model \\ \hline
\multicolumn{4}{c}{\textbf{\ref{TJ ip} Inverse dynamics methods}} \\ \hline
NeuroFluid~\cite{guan2022neurofluid} & \textit{ICML2022}  & MLP & Fluid dynamics grounding  \\ \hline
IFD~\cite{liu2023inferring} & \textit{Arxiv2023}  & CNN &  Fluid and beck dam particle dynamics inversion  \\ \hline
Deng \textit{et al.}~\cite{denglearning}  & \textit{ICLR2023}  & CNN & Inverse Navier–Stokes equations  \\ \hline
HyFluid~\cite{xu2025hybrid}   & \textit{Arxiv2025}  & GNN & Fluid dynamics grounding   \\ \hline
Zhu \textit{et al.}~\cite{zhulatent} & \textit{ICLR2024}  & CNN & Fluid dynamics grounding  \\ \hline
SciML~\cite{liudata} & \textit{3DV2026}  & Transformer &  Fluid dynamics grounding  \\ \hline
GIC~\cite{cai2024gic}  & \textit{NeurIPS2024}  & MLP & System identification from video\\ \hline
MASIV~\cite{zhao2025toward}& \textit{ICCV2025}  & MLP & System identification from video  \\ \hline

ProJo4D~\cite{rho2025projo4d}& \textit{Arxiv2025}  & MLP & System identification from video  \\ \hline
\multicolumn{4}{c}{\textbf{\ref{GB ip} Generative-based methods}} \\ \hline
\multicolumn{4}{l}{\textbf{\ref{db ip} Diffusion-based}} \\ \hline
Song \textit{et al.}~\cite{songsolving}& \textit{ICLR2022}  & UNet & Inverse problems for medical imaging  \\ \hline
DPS~\cite{chung2022diffusion}& \textit{ICLR2023}  & UNet & Inverse problems for low-level imaging  \\ \hline
MCG~\cite{chung2022improving}& \textit{NeurIPS2022}  & UNet & Inverse problems for low-level imaging  \\ \hline
Resample~\cite{song2023solving}& \textit{ICLR2024}  & UNet & Inverse problems for low-level imaging  \\ \hline
DiffusionPDE~\cite{huang2024diffusionpde}& \textit{NeurIPS2024}  & UNet & Inverse (Darcy Flow, Poisson, Helmholtz, Navier-Stokes) \\ \hline
 FunDPS~\cite{yao2025guided}& \textit{NeurIPS2025}  & UNet & Inverse (Darcy Flow, Poisson, Helmholtz, Navier-Stokes)
 \\ \hline
DDIS~\cite{lin2026decoupled}& \textit{ICLR2026}  & UNet & Inverse (Darcy Flow, Poisson, Helmholtz, Navier-Stokes) \\ \hline
FunDiff~\cite{wang2025fundiff}& \textit{Nature Communication2025}  & UNet & Sinusoidal functions, Kolmogorov flow, Burgers’ equation \\ \hline
\multicolumn{4}{l}{\textbf{\ref{gan ip} GAN-based}} \\ \hline
GANO~\cite{rahman2022generative}& \textit{TMLR2022}  & UNet & Volcanic activities 
Navier-Stokes \\ \hline
GAROM~\cite{coscia2024generative}& \textit{Scientific Reports2024}  & UNet & Reduced order modelling \\ \hline
\end{tabular}}
\vspace{-0.3cm}
\end{table*}


\section{Inverse Problems}\label{sec:IP}

\begin{wrapfigure}{r}{0.45\linewidth}
    \centering
    \includegraphics[width=\linewidth]{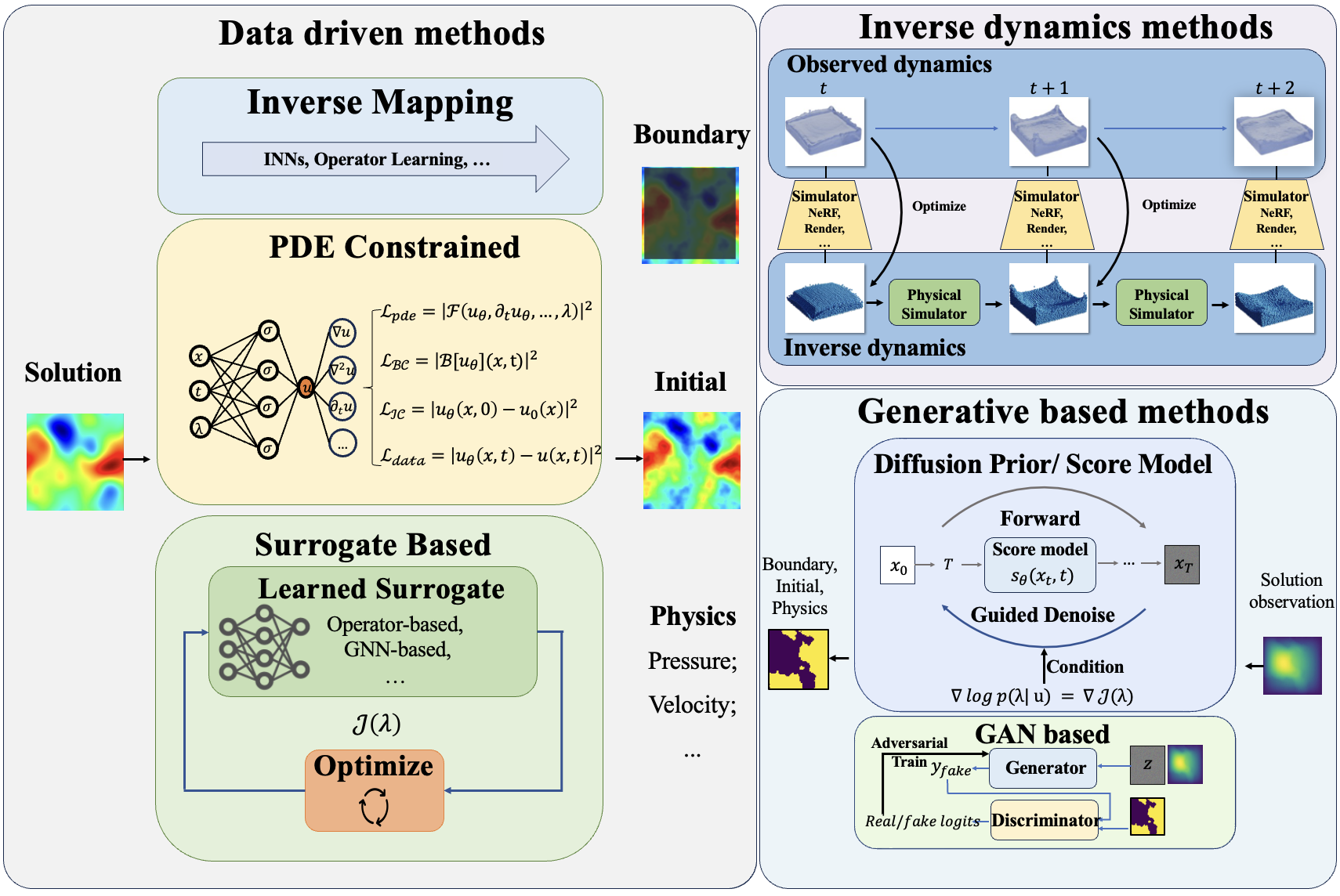}
    \caption{Overview of general AI methods for inverse problems, some simulated images are obtained from \cite{wang2025fundiff, wei2024diffphycon, li2024neuralfluid, wu2026geopt}}
    \label{fig:ipoverview}
\end{wrapfigure}
Inverse problems are fundamental to many scientific and engineering applications, where one seeks to recover an unknown signal, image, or parameter field from incomplete or indirect measurements. 

Representative examples include medical imaging modalities such as magnetic resonance imaging (MRI), computed tomography (CT)~\cite{songsolving}, and electrical impedance tomography (EIT)~\cite{denker2025deep}; full waveform inversion in geophysics~\cite{wu2019inversionnet}; fluid dynamics~\cite {cai2024gic}; and a variety of engineering applications~\cite{zhang2025artificial}. In these settings, the forward model is typically governed by a PDE system that maps the unknown parameter $\lambda$ to the corresponding state $u_\lambda$, with $\lambda$ and $u_\lambda$ defined as in Section~\ref{subsec:tradition}. Given the observation \(u_{\mathrm{obs}}\), one seeks to solve the following optimization problem with \(\mathcal{J}\) denoting the objective functional: 
\begin{equation}
\label{ip optimize}
        \lambda^*
=
\arg\min_{\lambda}
\mathcal{J}(u_\lambda,u_{\mathrm{obs}}).
\end{equation}


Inverse problems are ill-posed in several ways: (1) \textbf{non-existence}, where noisy observations or inaccurate forward models can make the inverse problem inconsistent; (2) \textbf{non-uniqueness}, where multiple solutions may correspond to the same observation, as in image inpainting; and (3) \textbf{instability}, where small perturbations in the observations can cause large changes in the recovered solution. To address these challenges, researchers have increasingly explored AI-based methods for inverse problems, motivated by advances in machine learning techniques, the expressive power of neural networks, and their success in PDE applications, which offer the potential to improve robustness, efficiency, and reconstruction quality. We organize this literature into three main categories: \textit{data-driven methods}, which directly learn a map from observations to unknown physical parameters; \textit{inverse dynamics methods}, which infer inverse dynamics under rollout-based observational constraints; and \textit{generative methods}, which model the distribution of physical parameters and generate them from Gaussian noise. Readers can refer to Fig. \ref{fig:ipoverview} for a clear understanding.

\textbf{Remark.} As inverse design and control problems share a similar optimization formulation with \eqref{ip optimize}, the categories above also extend naturally to these problems. Nonetheless, they introduce additional design-space and temporal-decision structures, which distinguish them slightly from inverse problems; therefore, they are discussed separately in Section \ref{sec:ID} and \ref{sec:IC}.

\subsection{Data-Driven Methods}
\label{dd ip}

General data-driven methods constitute the most fundamental class of AI approaches for inverse problems. With the rapid growth of data availability and computational resources, these methods have become increasingly effective at learning latent patterns from large-scale datasets to solve complex inverse tasks. In the following, we focus on three main categories: inverse-mapping, PDE-constrained, and surrogate-based methods.

\subsubsection{Inverse-Mapping}
\label{direct mapping ip}

Inverse-mapping methods are applicable not only to inverse problems governed by PDE systems, but also to broader inverse problems, as they directly learn the mapping from observations to the corresponding initial, boundary conditions, or unknown physics parameters. The formulation is given below:
\begin{equation}
    \lambda_\theta = f_\theta(u_{obs}),
\quad
\min_\theta
\mathbb{E}_{(u_{obs}, \lambda)}
\left[
\ell(f_\theta(u_{obs}),\lambda)
\right].
\end{equation}
Kendall \textit{et al.}~\cite{kendall2017uncertainties} first model $f_\theta$ with Bayesian neural network. However, Kruse \textit{et al.}~\cite{kruse2021benchmarking} noted that direct posterior learning requires a supervised loss that already reflects the posterior structure. When the loss fails to capture a complex posterior, such as a multimodal one, the learned solution can be inaccurate or misleading.

Another line of work, invertible neural networks (INNs)~\cite{ardizzoneanalyzing}, resolves the challenge by learning the forward process, which is much simpler, and then inverting it. INNs are particularly representative, as they enable estimation of the full posterior distribution for inverse problems. Important variants include conditional INNs (c-INNs)~\cite{kruse2021benchmarking}, i-ResNets~\cite{behrmann2019invertible}, invertible autoencoders~\cite{teng2019invertible}, and VINA~\cite{shekhar2022vina}. VINA further advances this direction by placing invertible models within a variational framework and providing theoretical guarantees for posterior approximation. Beyond strictly invertible architectures, other inverse-mapping methods have also been developed to characterize the ambiguity of inverse solutions. For example, mixture density network (MDN) based approaches~\cite{kruse2020technical,torfeh2025probabilistic} model the parameter space as a mixture of Gaussians, thereby providing a probabilistic description of multiple candidate solutions. 

Recent operator-learning methods have been increasingly adapted to inverse PDE problems. Neural Inverse Operators (NIOs) learn inverse maps with DeepONet- and FNO-style architectures~\cite{molinaro2023neural}, while iFNO introduces invertible Fourier blocks with variational latent variables to handle noise and ill-posedness~\cite{longinvertible}. LNO further improves efficiency by performing inversion in a latent space under sparse observations~\cite{wang2024latent}. DeepONet-based variants have also been extended to inverse PDEs on unknown point-cloud manifolds and to hidden-physics discovery from sparse data~\cite{jiao2026solving,sarkar2024learning}. More recently, IGNO~\cite{zang2025unified} introduces a unified physics-informed generative neural operator framework with a dual-encoder architecture, supporting both point measurements and operator-valued data with noise robustness

Overall, inverse-mapping methods offer an efficient and conceptually simple framework for inverse inference, although their effectiveness depends critically on how well they capture ambiguity and solution diversity.

\subsubsection{PDE-Constrained}
\label{pde constrained ip}

Among PDE-constrained methods, physics-informed neural networks (PINNs)~\cite{raissi2019physics} constitute a representative framework for inverse PDE problems. They jointly optimize the neural solution field and unknown physical parameters under PDE, initial, boundary, and data constraints, with the corresponding loss functions shown below:
\begin{equation}
    \min_{\theta,\lambda}\;
w_p \mathcal{L}_{\mathrm{PDE}}
+w_i \mathcal{L}_{\mathrm{IC}}
+w_b \mathcal{L}_{\mathrm{BC}}
+w_d \mathcal{L}_{\mathrm{data}},
\end{equation}
where $\theta$ denotes the neural network parameters and $\lambda$ follows the definition in Section \ref{subsec:tradition}. Here, $\mathcal{L}_{\mathrm{PDE}}$, $\mathcal{L}_{\mathrm{IC}}$, and $\mathcal{L}_{\mathrm{BC}}$ denote the losses associated with the PDE residual, initial conditions, and boundary conditions, respectively. The data loss, $\mathcal{L}_{\mathrm{data}}$, measures the discrepancy between the ground-truth observations and the predicted solution under the initial and boundary conditions together with the physical parameter $\lambda$. The corresponding loss terms are defined as follows, with $w_p$, $w_i$, $w_b$, and $w_d$ as the corresponding weights:

\begin{equation}
\begin{split}
    \mathcal{L}_{PDE} &= \frac{1}{|\Omega |} \sum_{(x, t) \in \Omega \times [0,T]} |\mathcal{F}(x,t,u_\theta,\nabla u_\theta,\nabla^2 u_\theta,\partial_t u_\theta,\ldots, \lambda_p)|^2,\\
    \mathcal{L}_{IC} &=\frac{1}{|\Omega|} \sum_{x \in \Omega} |u_{\theta}(x,0)-u_0(x)|^2,\\
\mathcal{L}_{BC} &=\frac{1}{|\partial \Omega|} \sum_{(x, t) \in \partial \Omega \times [0,T]} |\mathcal{B}[u_\theta](x,t)|^2, \\
    \mathcal{L}_{data} &=\frac{1}{|\Omega_o|} \sum_{(x, t) \in \Omega_o \times [0,T]} |u_\theta(x,t) - u_{obs}(x,t)|^2,
\end{split}
\end{equation}
where $\Omega_o$ denotes the observation set. 

Early work such as~\cite{chen2020physics} demonstrated the feasibility of PINNs for application-driven inverse problems in nano-optics, while gradient-enhanced PINNs (gPINNs)~\cite{yu2022gradient} improved inversion accuracy by enforcing derivatives of the PDE residual. Later, DG-PINNs~\cite{zhou2024data} and H-PINN~\cite{chandrasukmana2025hadamard} emphasized optimization strategy through staged training to address loss imbalance and poor initialization. Architectural improvements were further explored in~\cite{zhang2025inverse}, which introduced adaptive activation design to enhance convergence and accuracy.

Beyond strong-form residual minimization, alternative PDE formulations have also been used to construct network training objectives. In particular, E \textit{et al.}~\cite{E_2018} proposed the deep Ritz method, which employs a variational formulation for PDEs that can be written as energy minimization problems. Meanwhile, weak-form learning offers another important route. Zang~\textit{et al.}~\cite{zang2020weak} proposed weak adversarial networks (WANs), providing a weak-form, minimax-based framework for PDE learning with relaxed regularity requirements. Bao~\textit{et al.}~\cite{bao2020numerical} subsequently extended WANs to inverse problems such as electrical impedance tomography and dynamic electrical impedance tomography. More recently, ParticleWNN~\cite{Bao_2025} further improved efficiency and accuracy by adopting compactly supported radial basis functions as test functions. Following this weak-form perspective, WF-PINNs~\cite{wang2025wf} adopted weak-form residuals for problems with steep gradients, PIED~\cite{hemachandrapied} extended PINNs to inverse experimental design, and Perez \textit{et al.}~\cite{perez2025physics} generalized the framework to infinite and semi-infinite domains with Kolmogorov-Arnold networks (KAN)~\cite{liu2024kan}.

These studies show that PDE-constrained learning methods encompass a diverse range of categories and techniques, including strong-form PINN residual minimization, weak-form formulations, adversarial approaches, gradient-enhanced methods, and architecture-adaptive designs. However, PDE-constrained methods still face several core challenges: the composite loss is difficult to optimize jointly, automatic differentiation introduces substantial memory and computational overhead, and training performance is highly sensitive to the sampling of collocation and observations. To address these challenges, surrogate-based optimization methods have emerged.

\subsubsection{Surrogate-Based}
\label{surrogate ip}

Inverse-mapping and PDE-constrained methods typically do not explicitly exploit a learned forward simulator~\cite{yueholistic,yue2026deltaphi} when solving inverse problems. In contrast, surrogate-based methods first train a deep neural network to approximate the forward process, which serves as a computationally efficient substitute for potentially expensive numerical simulators~\cite{wu2024transolver,luotransolver++,ren2020benchmarking}. The learned surrogate model is then used as an efficient forward operator for inverse optimization and parameter recovery following equation \eqref{inverse optimize}.

An early representative of this paradigm is the neural-adjoint framework, which replaces the expensive forward simulator with a differentiable neural surrogate and recovers unknown parameters through gradient-based optimization~\cite{ren2020benchmarking}. Subsequent work extended this idea to PDE-constrained and irregular-domain settings, where graph-based surrogates enable efficient inversion from sparse observations~\cite{zhao2022learning}. More recent studies emphasize that surrogates should be evaluated not only by forward accuracy, but also by the inverse landscapes they induce. Examples include differentiable surrogates with physics-based regularization for bathymetry recovery~\cite{liu2024bathymetry}, PETAL with physics-based linearization for improved inversion~\cite{jin2023petal}, and hybrid methods that use surrogates as priors or initializations for downstream optimization or sampling~\cite{weidner2024learnable}.


Another line of work, known as operator-learning methods, was also developed for forward PDE surrogates. Compared with general surrogate models, these methods better handle multiscale structure and varying resolutions. DeepONet~\cite{lu2019deeponet} introduced a branch--trunk architecture for nonlinear operator approximation, FNO~\cite{li2020fourier} improved scalability through Fourier-domain convolutions, and Geo-FNO~\cite{li2023fourier} extended this framework to irregular geometries via learned domain deformation. Together, these works established operator learning as a function-space framework for inverse problems, where one aims to recover unknown fields or coefficients from partial and indirect observations.


\subsection{Inverse Dynamics Methods}
\label{TJ ip}

Inverse dynamics methods recover hidden time-evolving physical states from visual observations by coupling two trajectories over time.
Therefore, the inverse problem is solved not by a one-shot mapping from observations to parameters, but by iteratively aligning the latent physical trajectory so that its simulated and rendered rollout matches the observed visual trajectory. A generic formulation is
\begin{equation}
\begin{split}
    z_{t+1} = f_{\theta}(z_t,\lambda), \quad
\hat{u}_t = Render_{\phi}(z_t), \\
\min_{\theta,\phi,z_{0:T},\lambda} \quad 
\sum_{t=1}^{T}\mathcal{L}(\hat{u}_t,u_t)
+\mathcal{R}_{inv}(z_{0:T},\lambda),
\end{split}
\end{equation}
where $u_{1:T}$ is the observed visual trajectory, $z_{0:T}$ is the hidden physical trajectory, $\lambda$ denotes the intrinsic unknown physical parameters, $f_{\theta}$ is a differentiable dynamics model, $\mathcal{R}_{inv}$ is the regularization term, and $Render_{\phi}$ is the rendering or observation operator. 


Early instances of this paradigm appear in differentiable simulation--rendering frameworks for fluid dynamics. NeuroFluid~\cite{guan2022neurofluid} jointly optimizes over a particle transition model and a particle-based neural renderer, 
Similarly, IFD~\cite{liu2023inferring} couples a differentiable Euler simulator with a volumetric renderer. 
Subsequent work imposes more structure on the latent trajectory itself. For example, \cite{denglearning} replaces dense fluid fields with low-dimensional differentiable vortex particles.
HyFluid~\cite{xu2025hybrid} further strengthens the coupling by jointly inferring density and velocity fields from sparse-view videos.

Another important development is that the hidden trajectory need not be parameterized as explicit material coefficients. Zhu \textit{et al.}~\cite{zhulatent} models hidden physics through probabilistic latent states that approximate the visual posterior, and then transfers the inferred dynamics to novel scenes. SciML~\cite{liudata} extends this line from a data-efficiency perspective by using a foundation model to forecast augmented frames and distill feature representations back into the inverse model. Similarly, Zhao \textit{et al.}~\cite{zhao20253d} pushes this paradigm into monocular settings by reconstructing 3D dynamic fluid assets from single-view videos, estimating surface motion, lifting it to volumetric velocity, and optimizing simulation parameters across time.

More recent work extends this paradigm beyond fluids to broader inverse physics and system identification. GIC~\cite{cai2024gic} couples observed deformation trajectories with simulated continuum dynamics through dynamic 3D Gaussians, reconstructed continuums, and differentiable mask rendering for physical property identification. MASIV~\cite{zhao2025toward} further relaxes predefined material priors by using reconstructed dense particle trajectories to guide the learning of neural constitutive laws, extending the coupling from state trajectories to governing dynamics. ProJo4D~\cite{rho2025projo4d} addresses sparse-view inverse physics by expanding the jointly optimized variable set.


The common principle of these methods is not a specific simulator, renderer, or representation, but an inverse-dynamics coupling between observation space and latent physical space. Compared with direct inverse mappings, they often improve temporal consistency and physical plausibility, but they also introduce highly nonconvex optimization, in which rendering–dynamics–parameter coupling can lead to ambiguity and error accumulation. Recent advances therefore emphasize better latent parameterizations, stronger temporal or geometric supervision, and more stable optimization. Overall, trajectory-coupled inverse dynamics provides a useful lens for methods that jointly infer observed motion and its underlying physical evolution.


\subsection{Generative-Based Methods}
\label{GB ip}

Generative models~\cite{goodfellow2014generative,ho2020denoising,song2020denoising,nichol2021improved,song2020score,rombach2022high} have become increasingly prominent due to their remarkable performance in visual generation tasks. In essence, these models learn a probability distribution over clean data and generate samples by transforming noise variables
into structured outputs. From this perspective, applying generative models to inverse problems can be viewed as a conditional generation task. Given observations $u_{obs}$ and a generative prior $p_\theta(\cdot | \mathcal{F})$ learned from the PDE system $\mathcal{F}$, the goal is to infer the corresponding initial condition or physical parameter $\lambda$ by modeling the conditional distribution $p_{\theta}(\lambda| u_{obs},\mathcal{F})$. In the following sections, we review these methods from two main perspectives: diffusion-based methods and generative adversarial network (GAN)-based methods.

\subsubsection{Diffusion-Based}
\label{db ip}

Diffusion models~\cite{ho2020denoising,song2020score} are generative models that learn the data distribution by reversing a progressive noising process. In continuous time, it can be formulated as the SDE~\cite{song2020score}
\begin{equation}
\label{eq:reverse_sde}
\mathrm{d}x
=
\left[
f(x,t)-g(t)^2\nabla_x \log p_t(x)
\right]\mathrm{d}\bar{t}
+
g(t)\,\mathrm{d}\bar{w},
\end{equation}
where $w$ is a Wiener process, $f$ is the drift, and $g$ is the diffusion coefficient. $\bar{t}$ denotes reverse time and $\nabla_x \log p_t(x)$ is approximated by a score function. For conditional generation, by Bayes' rule, the score in \eqref{eq:reverse_sde} becomes
\begin{equation}
\label{condition}
\nabla_x \log p_t(x| y)
=
\nabla_x \log p_t(x)
+
\nabla_x \log p_t(y| x).
\end{equation}

In inverse problems, the condition is the observation \(y=u_{obs}\), while the generated variables are the initial, boundary condition, or unknown physical parameters \(\lambda = (\lambda_p,u_0,\mathcal{B})\) and possibly the PDE inputs \(x,t\). Diffusion-based approaches typically first train a score model to capture the distribution of the physical parameters and then perform conditional sampling via \eqref{condition} to infer the specific physical parameters \(\lambda\). Specifically $p_t(y| x) = p_t(u|x,t,\lambda)$ is usually characterized as:
\begin{equation}
   \log p(u_{obs}|\lambda)
\propto
-\frac{1}{2\sigma^2}
\|u_\lambda-u_{obs}\|^2.
\end{equation}

Building upon this foundational idea, a broad family of diffusion-based sampling methods has emerged for solving inverse problems not only under PDE scenarios. In particular, Song \textit{et al.}~\cite{songsolving} first showed that a score-based generative model can recover image signal through condition sampling follows equation \eqref{condition}. Building on this perspective, DPS~\cite{chung2022diffusion} further generalized the framework to nonlinear inverse problems by approximate posterior sampling. Since then, many subsequent methods~\cite{yu2023freedom,bansal2023universal,song2023loss,chung2022improving,he2023manifold,efron2011tweedie,song2023solving,li2024decoupled,zhu2023denoising,ye2024tfg} have followed equation~\eqref{condition} and achieved strong performance across a wide range of applications. A systematic taxonomy of these methods is provided in~\cite{daras2024survey}.

For PDE-governed inverse problems, recent diffusion-based methods extend posterior sampling from image space to function space. DiffusionPDE models the joint distribution of PDE coefficients and solutions, supporting missing-data completion as well as forward and inverse solving under partial observations~\cite{huang2024diffusionpde}. FunDPS further introduces a discretization-agnostic function-space diffusion model with plug-and-play guidance for sparse or noisy measurements~\cite{yao2025guided}. More recent methods adopt a decoupled design: DDIS learns an unconditional diffusion prior over coefficient fields and uses a neural operator for PDE-guided sampling~\cite{lin2026decoupled}, while Fun-DDPS applies a similar strategy to subsurface CO$_2$ modeling and data assimilation~\cite{ju2026function}. Related works such as FunDiff and diffusion-prior ensemble samplers further emphasize physics-informed function-space generation and Bayesian posterior sampling~\cite{wang2025fundiff,chen2025solving}. Overall, this line of work marks a shift from image-space heuristic guidance toward function-space, physics-aware, and decoupled diffusion frameworks for inverse PDE problems.


\subsubsection{GAN-Based}
\label{gan ip}

Compared with diffusion-based methods, GAN-based approaches offer an earlier and more amortized generative paradigm for inverse problems. Instead of iterative posterior refinement at inference time, they learn the solution manifold through adversarial training, enabling inversion via conditional generation after training. In PDE-related settings, GAN-based methods has mainly appeared in two forms: GANO~\cite{rahman2022generative} extends GANs to infinite-dimensional function spaces by combining a neural-operator generator with a functional discriminator, while GAROM~\cite{coscia2024generative} introduces a conditional adversarial reduced-order model that generates high-fidelity solution fields from PDE parameters and supports uncertainty-aware prediction.

From the perspective of inverse problems, these GAN-based methods are appealing because they replace expensive iterative optimization or sampling with fast feed-forward generation after training. However, compared with diffusion-based posterior sampling methods, GAN-based approaches usually provide weaker coverage of multi-modal posteriors and are more prone to training instability or mode collapse, which may limit their robustness in severely ill-posed inverse settings.

\section{Inverse Design}\label{sec:ID}

\begin{table*}[ht]
\centering
\scriptsize
\caption{Overview of AI methods for inverse design.}
\label{tab:id}
\vspace{-0.2cm}

\setlength{\tabcolsep}{4.5mm}{
\begin{tabular}{l|c|c|c}
\hline
\textbf{Methods} & \textbf{Source} & \textbf{Backbone} & \textbf{Tasks} \\ \hline
\multicolumn{4}{c}{\textbf{\ref{ID:Opt} Optimization-based inverse design}} \\ \hline
\multicolumn{4}{l}{\textbf{\ref{ID:Opt:PhysicsLoop} Physics-in-the-loop}} \\ \hline
Luce \textit{et al.}~\cite{luce2024merging} & \textit{MLST2024} & Differentiable Solver & Photonic inverse design \\ \hline
JAX-Fluids ~\cite{bezgin2023jax} & \textit{CPC2023} & Differentiable Solver & Fluid inverse design \\ \hline
Neuralfluid~\cite{li2024neuralfluid} & \textit{NeurIPS2024} & Differentiable Solver & Fluid inverse design \\ \hline
$\phi_{flow}$~\cite{holl2024bf} & \textit{ICML2024} & Differentiable Solver & Fluid inverse design \\ \hline
Chen \textit{et al.}~\cite{chen2020physics} & \textit{Opt. Express2020} & MLP & Inverse scattering design \\ \hline
hPINNs~\cite{lu2021physics} & \textit{SIAM SISC2021} & MLP & Optics and Fluid inverse design \\ \hline
Zhao \textit{et al.}~\cite{zhao2025physics} & \textit{Neural Networks2025} & MLP & Temperature field/Flow field inverse design \\ \hline
Hao \textit{et al.}~\cite{hao2022bi} & \textit{ICLR2023} & MLP & Fluid inverse design \\ \hline

\multicolumn{4}{l}{\textbf{\ref{ID:Opt:surrogate} Learned-model-assisted}} \\ \hline
Allen \textit{et al.}~\cite{allen2022inverse} & \textit{NeurIPS2022} & GNN & Fluid-structure inverse design \\ \hline
Shukla \textit{et al.}~\cite{shukla2024deep} & \textit{EAAI2024} & MLP & Airfoil inverse design \\ \hline
Lu \textit{et al.}~\cite{lu2022multifidelity} & \textit{PRR2022} & MLP & Nanoscale heat transport design \\ \hline
Kumar \textit{et al.}~\cite{kumar2020inverse} & \textit{npj Comput. Mater.2020} & MLP & Spinodoid metamaterial design \\ \hline
Augenstein \textit{et al.}~\cite{augenstein2023neural} & \textit{ACS Photonics2023} & CNN & Photonic inverse design \\ \hline

\multicolumn{4}{c}{\textbf{\ref{ID:Dir} Direct inverse design}} \\ \hline
Liu \textit{et al.}~\cite{liu2018training} & \textit{ACS Photonics2018} & MLP & Nanophotonic inverse design \\ \hline
Tahersima \textit{et al.}~\cite{tahersima2019deep} & \textit{Sci. Rep.2019} & MLP & Photonic power splitter design \\ \hline
Malkiel \textit{et al.}~\cite{malkiel2018plasmonic} & \textit{Light Sci. Appl.2018} & MLP & Plasmonic nanostructure design \\ \hline
An \textit{et al.}~\cite{an2019deep} & \textit{ACS Photonics 2019} & MLP & All-dielectric metasurface design \\ \hline
Peurifoy \textit{et al.}~\cite{peurifoy2018nanophotonic} & \textit{Sci. Adv. 2018} & MLP & Nanophotonic particle design \\ \hline
Unni \textit{et al.}~\cite{unni2021mixture} & \textit{Nanophotonics2021} & MLP & Thin-film reflector design \\ \hline
AutoTandemML ~\cite{grbcic2025autotandemml} & \textit{Appl. Soft Comput.2025} & MLP & Data-efficient inverse design \\ \hline
Grbcic \textit{et al.}~\cite{grbvcic2025inverse} & \textit{Comput. Mater. Sci.2025} & MLP &  Photonic surfaces inverse design \\ \hline
Ma \textit{et al.}~\cite{ma2018deep} & \textit{ACS NANO2018} & CNN & Metamaterial design \\ \hline
Sekar \textit{et al.}~\cite{sekar2019inverse} & \textit{AIAA J.2019} & CNN & Airfoil inverse design \\ \hline

\multicolumn{4}{c}{\textbf{\ref{ID:Gen} Generative inverse design}} \\ \hline
\multicolumn{4}{l}{\textbf{\ref{ID:Gen:GAN} Latent-variable and adversarial}} \\ \hline
Liu \textit{et al.}~\cite{liu2018generative} & \textit{Nano Lett.2018} & GAN & Metasurface inverse design \\ \hline
Oh \textit{et al.}~\cite{oh2019deep} & \textit{J. Mech. Des.2019} & GAN & Topology inverse design \\ \hline
Achour \textit{et al.}~\cite{achour2020development} & \textit{AIAA2020} & GAN & Airfoil inverse design \\ \hline
PcDGAN ~\cite{heyrani2021pcdgan} & \textit{SIGKDD} & GAN & Airfoil inverse design \\ \hline
GLOnet~\cite{jiang2020simulator} & \textit{Nanophotonics2020} & GAN & Metasurface inverse design \\ \hline
Wang \textit{et al.}~\cite{wang2020deep} & \textit{CMAME2020} & VAE & Topology inverse design \\ \hline
Yang \textit{et al.}~\cite{yang2023inverse} & \textit{Eng. Comput.2023} & VAE & Wind turbine airfoil inverse design \\ \hline
A-CVAE~\cite{tang2020generative} & \textit{Laser Photonics Rev.2020} & VAE & Nanophotonic device design \\ \hline

\multicolumn{4}{l}{\textbf{\ref{ID:Gen:Diffusion} Diffusion-based}} \\ \hline
Cindm ~\cite{wu2024compositional} & \textit{ICLR2023} & Diffusion & Multi-body/Airfoil design \\ \hline
ZeoDiff~\cite{park2024inverse} & \textit{J. Mater. Chem.2024} & Diffusion& Porous material discovery \\ \hline
TopoDiff~\cite{maze2023diffusion} & \textit{AAAI2023} & Diffusion & Topology optimization \\ \hline
MetaGen~\cite{hen2025inverse} & \textit{ASC Photonics2025} & Diffusion& Photonic inverse design \\ \hline
PhysGen~\cite{you2025physgen} & \textit{CVPR2026} & Diffusion & 3D physics-guided inverse design \\ \hline
3DID~\cite{hao20253did} & \textit{NeurIPS2025} & Diffusion & Aerodynamic inverse design \\ \hline
Bastek \textit{et al.}~\cite{bastek2023inverse} & \textit{Nat. Mach. Intell.2023} & Diffusion & Mechanical metamaterial design \\ \hline
Yang \textit{et al.}~\cite{yang2024guided} & \textit{Arxiv2024} & Diffusion & Mechanical metamaterial inverse design \\ \hline
\end{tabular}}
\vspace{-0.3cm}
\end{table*}

As a specific branch of broader inverse problems, inverse design shifts the focus from recovering unknown physical quantities from observations to synthesizing design variables that realize a desired physical behavior. More specifically, the aim of inverse design is to identify a set of design variables, such as geometry, topology, material distribution, or structural parameters, that produce a target PDE-governed response. This problem appears in a wide range of scientific and engineering applications, including mechanical systems~\cite{bastek2023inverse, zeni2025generative}, aerodynamic problems~\cite{song2023surrogate}, and physics detector development~\cite{schuetz2024resum}. In PDE-governed settings, inverse design aims to identify a design variable $d$ such that the resulting physical response satisfies a prescribed objective. Here, $u(d)$ denotes the physical field induced by $d$ through the corresponding PDE system. We define the design objective as: 
\begin{equation}
\mathcal{J}(d)\coloneqq\mathcal{J}\big(d,u(d)\big),
\end{equation}
where the dependence on $d$ arises both through the induced physical response $u(d)$ and through the properties of the design itself. A classical inverse design problem is therefore formulated as: 
\begin{equation}
\begin{split}
d^\ast &= \arg\min_d \mathcal{J}\big(d,u(d)\big) \\
&\text{s.t.}\quad \mathcal{C}\big(d,u(d)\big)\le 0,
\end{split}
\end{equation}

where the solution $d^\ast$ denotes the design that minimizes the performance objective, and $\mathcal{C}$ aggregates problem-dependent design constraints, such as volume, manufacturability, and bound constraints.

\noindent Compared with general inverse problems, inverse design seeks to actively create structures or configurations with prescribed functionality. This shift from estimation to synthesis also brings distinctive challenges: 
(1)~\textbf{Large and structured design spaces}: unlike parameter estimation, where unknowns are typically parameter fields of fixed dimension, the design variable $d$ in inverse design often spans large discrete spaces (e.g., topologies and material selections) or high-dimensional continuous manifolds, making the search space much larger and the optimization process substantially more challenging.

(2)~\textbf{Non-convex optimization landscapes}: the objective $\mathcal{J}\big(d,u(d)\big)$ in inverse design is typically highly non-convex, since small changes in the design may induce complicated and nonlinear changes in the resulting physical response. As a result, the optimization landscape often contains many local minima and flat regions, which greatly increases the difficulty of design search.
(3)~\textbf{Expensive PDE-constrained evaluation}: evaluating $\mathcal{J}\big(d,u(d)\big)$ requires computing the induced physical field $u(d)$ by solving the corresponding forward PDE system. As a result, both objective evaluation and gradient-based design updates can become computationally expensive, making large-scale design exploration and optimization prohibitively costly.

These challenges motivate various strategies for producing candidate designs. Accordingly, we organize the existing literature into three categories: \textit{optimization-based inverse design}, which iteratively refines $d$
using gradients, adjoint information, or learned surrogates; \textit{direct inverse design}, which learns a direct mapping from target response to design; \textit{generative inverse design}, which explicitly models the distribution of feasible designs to address one-to-many solutions. Finally, we provide a dedicated discussion on \textit{design space representation}, a problem-specific supporting dimension of inverse design.

\subsection{Optimization-Based Inverse Design}
\label{ID:Opt}
Optimization-based inverse design treats inverse design as an iterative optimization problem, in which the design variable $d$ is progressively updated to minimize a task-specific objective associated with a PDE-governed system. In practice, such optimization can be carried out either by keeping the governing physics directly in the loop or by replacing expensive PDE evaluations with learned forward models. Accordingly, we categorize existing approaches into two classes: \textit{physics-in-the-loop methods}, which keep the governing PDE directly involved in optimization through differentiable solvers or physics-residual-based objectives; and \textit{learned-model-assisted methods}, which accelerate design optimization by using learned surrogates or neural operators.

\subsubsection{Physics-in-the-Loop Inverse Design}
\label{ID:Opt:PhysicsLoop}
\begin{wrapfigure}{r}{0.45\linewidth}
    \centering
    \includegraphics[width=\linewidth]{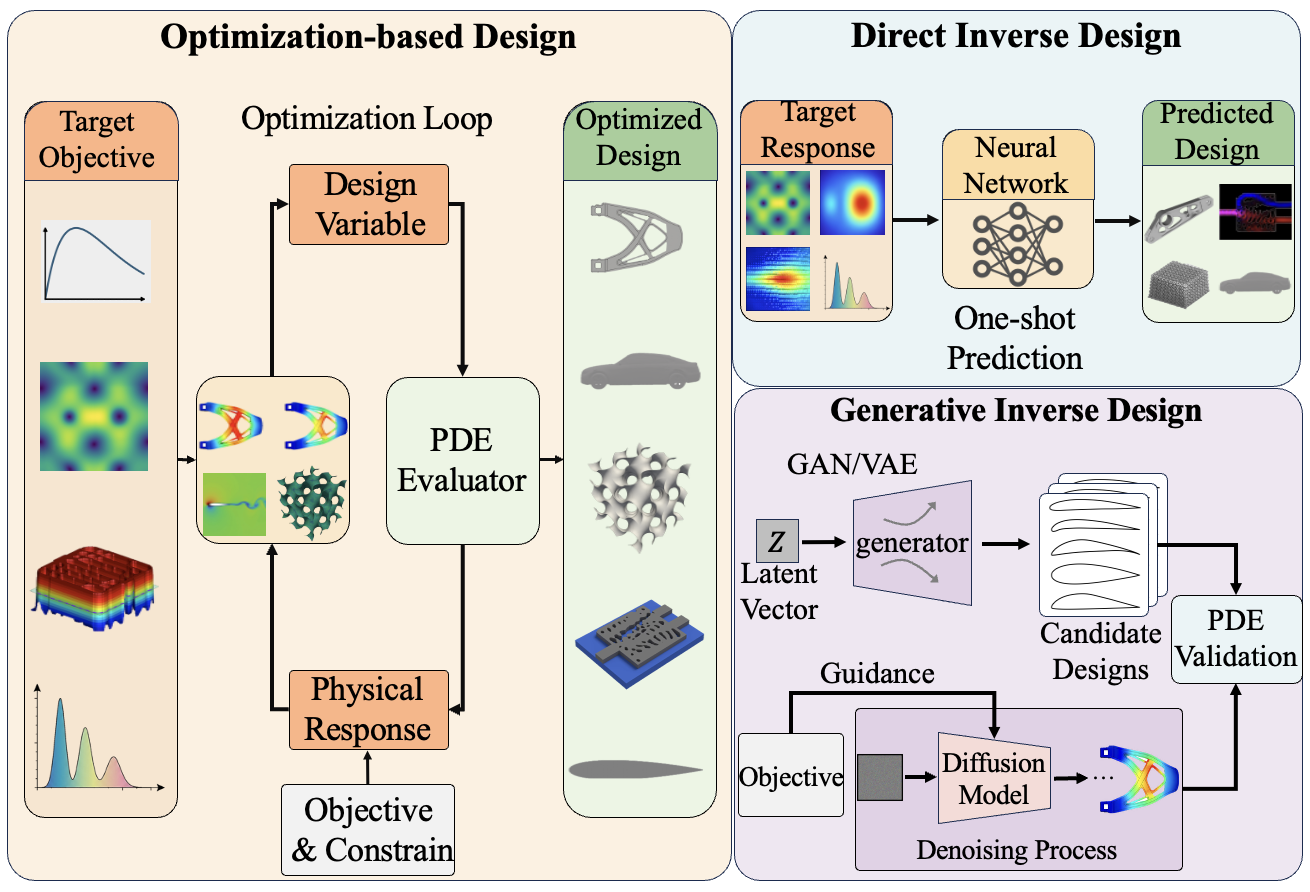} 
    \caption{Overview of general AI methods for inverse design.}
    \label{fig:IDoverview}
\end{wrapfigure}

Physics-in-the-loop inverse design keeps the governing PDE directly in the optimization loop and updates the design variable by leveraging gradients derived either from differentiable solvers or from physics-residual-based objectives. These methods maintain tight coupling between design optimization and the underlying PDE, thereby preserving strong physical consistency during the search process.

One representative line of work makes the full forward PDE process differentiable with respect to the design variables, enabling end-to-end gradient-based optimization with modern automatic differentiation tools~\cite{paszke2019pytorch, abadi2016tensorflow, bradbury2018jax}. Representative examples include~\cite{luce2024merging}, which combines automatic differentiation with adjoint-based formulations for photonic inverse design. In fluid settings,~\cite{bezgin2023jax} demonstrates that high-order computational fluid dynamics (CFD) solvers themselves can be made differentiable, opening the door to gradient-based inverse design in more complex flow regimes. Recently,~\cite{li2024neuralfluid} further illustrates the potential of differentiable PDE solvers for design in fluid systems. Related advances in solver-in-the-loop differentiable physics and software infrastructure, such as~\cite{holl2024bf}, further broaden the applicability of this paradigm beyond specific inverse design tasks.

Another line incorporates the governing PDE directly into the optimization objective through physics-informed residual constraints. Building on the foundational work of PINNs~\cite{raissi2019physics}, ~\cite{lu2021physics} extends this framework to optics and fluids problems via hard constraint enforcement, while Chen \textit{et al.}~\cite{chen2020physics} extend it to inverse scattering and broader nanophotonic design problems. The same physics-residual idea has proven effective across diverse systems, including metamaterial design~\cite{chen2025physics} 
and inverse design in phase field models~\cite{zhao2025physics}. A persistent challenge across these settings is multi-term loss instability, where PDE residual and design objective terms compete during optimization. Structural solutions include the bilevel formulations of~\cite{hao2022bi} and~\cite{zhang2024bilo}, which decouple the design objective from the PDE constraint and thereby reduce interference between the two optimization levels. Alternatively,~\cite{Bao_2026} integrates WAN~\cite{zang2020weak} with the augmented Lagrangian method for interface optimal design, yielding improved robustness, high-dimensional applicability, and better avoidance of poor local minima.

Taken together, physics-in-the-loop methods preserve strong physical fidelity by keeping the governing PDE directly in the optimization loop, but they remain computationally expensive and can be sensitive to optimization instability, especially in large-scale design spaces.

\subsubsection{Learned-Model-Assisted Inverse Design}
\label{ID:Opt:surrogate}
Learned-model-assisted inverse design accelerates optimization by replacing expensive PDE evaluations with learned forward models, thereby substantially reducing the per-iteration cost and making large-scale design exploration more practical~\cite{azizzadenesheli2024neural,kumar2020inverse}.

Early examples of this idea include~\cite{allen2022inverse}, which uses learned simulators to accelerate inverse design in fluid-structure systems, and~\cite{shukla2024deep}, which directly leverages neural operators for optimization-oriented forward prediction. ~\cite{lu2022multifidelity} further shows that multifidelity operator learning can improve the efficiency-accuracy trade-off in design tasks involving expensive PDE evaluations. More recently,~\cite{cheng2025accelerating} demonstrates that neural operators can facilitate PDE-constrained optimization by enhancing derivative learning and promoting faster convergence.

Beyond these representative examples, neural operators are increasingly adopted in more specialized design settings. ~\cite{augenstein2023neural} develops a neural-operator surrogate for free-form photonic device design, while~\cite{erzmann2024equivariant} uses neural operators as PDE surrogates within topology optimization. Meanwhile,~\cite{li2022empowering} provides a broader overview of applications for metasurfaces design.

Compared with physics-in-the-loop approaches, learned-model-assisted methods trade some degree of physical fidelity for much higher computational efficiency. They are therefore particularly attractive when repeated PDE evaluation is the main bottleneck and when a sufficiently accurate forward surrogate can be learned in advance.

\subsection{Direct Inverse Design}
\label{ID:Dir}
Direct inverse design learns a direct mapping from target responses to design variables, so that a feasible design can be generated in a single forward pass without iterative optimization. A representative example is~\cite{liu2018training}, which introduces a tandem architecture to alleviate the non-uniqueness of the inverse map. Similar ideas are applied across a range of applications, including photonic power splitters~\cite{tahersima2019deep}, aerodynamic shape design~\cite{sekar2019inverse}, nanophotonic particle design~\cite{peurifoy2018nanophotonic}, plasmonic nanostructures~\cite{malkiel2018plasmonic}, all-dielectric metasurfaces~\cite{an2019deep}, and chiral metamaterials~\cite{ma2018deep}. More recent extensions, such as~\cite{unni2021mixture,grbcic2025autotandemml,grbvcic2025inverse}, further improve this paradigm by addressing ambiguity and improving data efficiency. Overall, direct inverse design provides a simple and efficient route to one-shot design synthesis, although its deterministic nature often limits its ability to fully capture the one-to-many design space.

\subsection{Generative Inverse Design}
\label{ID:Gen}
Generative inverse design addresses inverse design by learning a prior over feasible designs and generating candidates conditioned on target responses, rather than searching for solutions through iterative PDE-constrained optimization. This formulation is particularly suitable for PDE-governed design, where the response-to-design map is often inherently one-to-many and multiple distinct designs may satisfy the same target behavior. Instead of collapsing these possibilities into a single prediction, generative models aim to capture the underlying design distribution and produce diverse feasible candidates. According to the underlying generative paradigm, we categorize existing methods into two groups: latent-variable and adversarial generative models, and diffusion-based inverse design.

\subsubsection{Latent-Variable and Adversarial Generative Models}
\label{ID:Gen:GAN}
In the early stage of generative inverse design, adversarial and latent-variable models serve as the main paradigms for addressing the one-to-many nature of design synthesis. Among them, GAN-based~\cite{goodfellow2014generative} methods provide an early generative route by learning the design manifold through adversarial training and then producing candidate designs via fast conditional generation, rather than solving a new PDE-constrained optimization problem for each target. In PDE-governed settings, this idea has been explored in a range of applications, including metasurface design~\cite{liu2018generative}, structural topology generation~\cite{oh2019deep}, aerodynamic shape generation~\cite{achour2020development,heyrani2021pcdgan}, and metamaterial synthesis~\cite{wang2020deep}. These methods are attractive because they replace expensive iterative optimization with efficient feed-forward design generation after training, although their ability to capture highly multimodal design distributions is often limited by training instability and mode collapse.

By contrast, latent-variable methods, especially variational autoencoders (VAEs)~\cite{kingma2013auto}, provide a more structured probabilistic paradigm for generative inverse design. Instead of learning the design manifold purely through adversarial training, these methods construct a compact latent space over feasible designs, so that inverse design can be carried out through conditional sampling, latent-space optimization, and uncertainty-aware search. Representative examples include~\cite{yang2023inverse}, which adopts a latent-variable model for wind turbine airfoil generation, and conditional VAE-based frameworks for integrated nanophotonic device design~\cite{tang2020generative}. Related latent-space strategies are also explored for free-form photonic design and metasurface generation~\cite{marzban2026inverse,marzban2025hilab}. Overall, these methods offer more controllable representations for design exploration and refinement, although their effectiveness still depends on whether the learned latent space can adequately capture design diversity, physical validity, and the non-uniqueness of the underlying design space.

\subsubsection{Diffusion-Based Inverse Design}
\label{ID:Gen:Diffusion}
Among existing generative paradigms, diffusion models have recently emerged as one of the most effective approaches for inverse design, especially in topology-rich and high-dimensional settings. Early representative works have demonstrated the effectiveness of diffusion models across a range of inverse design tasks. In mechanics,~\cite{bastek2023inverse} shows that diffusion models can generate diverse nonlinear mechanical metamaterial designs conditioned on target stress, while~\cite{yang2024guided} further highlights their efficiency in density-based mechanical metamaterial design. Similar ideas were later extended to porous material discovery \cite{park2024inverse}, topology optimization \cite{maze2023diffusion}, and photonics \cite{hen2025inverse}. A common idea behind these methods is to cast inverse design as a conditional sampling problem. Similar to diffusion-based inverse problems discussed in Section~\ref{db ip}, a generative prior over feasible designs is learned first, and the target response is then introduced as guidance during reverse-time generation. In this way, diffusion models avoid collapsing the inverse map into a single deterministic prediction and instead support the generation of multiple candidate designs consistent with the desired objective.

Furthermore, inverse design poses challenges beyond those of standard inverse problems: the design space is often larger, more structured and more compositional, whereas valid solutions must satisfy not only target responses but also geometric plausibility and practical constraints. This has motivated recent work to move beyond standard conditional diffusion toward more structured design generation. For example, ~\cite{wu2024compositional} interprets diffusion models from an energy-based perspective, enabling compositional guidance during sampling. Related efforts in 3D inverse design, such as~\cite{you2025physgen} and~\cite{hao20253did}, further combine structured latent representations with physical guidance to support design in larger and more expressive geometric spaces. Taken together, these developments reflect a broader shift in inverse design: rather than using diffusion models only for standard target-conditioned sampling, recent work increasingly explores how their generative power can support more complex, structured, and physically grounded design tasks.

\subsection{Design Space Representation}
Beyond the optimization pipeline, inverse design also depends critically on how the design space is represented. Unlike general inverse problems, where the unknowns are often fixed physical parameters or fields, inverse design must specify the form of the design variable itself, which may correspond to geometric parameters, topology, or density fields, material distributions, graphs, or learned latent codes. This choice directly affects the expressiveness of the design space, the difficulty of optimization, and the ability to capture diverse feasible solutions.

Existing methods address the representation problem through several main strategies. A common choice is parametric representation~\cite{cummings2015applied,liu2024afbench, masters2017geometric,colburn2021inverse}, where designs are described by a fixed set of geometric or physical parameters, leading to a compact and optimization-friendly design space but limited expressiveness. A second line adopts field-based or topology-based representations, such as density fields or spatially distributed material variables~\cite{bendsoe1988generating, allaire2004structural, maze2023diffusion}. These representations offer greater flexibility for free-form design, but substantially enlarge the search space. More recently, learned representations, including latent spaces and neural implicit representations, have been introduced to encode complex design manifolds into more structured spaces for downstream optimization and generation~\cite{tang2020generative,park2019deepsdf,zehnder2021ntopo,zhang2023topology,hao20253did}. Taken together, design space representation serves as a fundamental supporting dimension of inverse design, since different representations offer different trade-offs between compactness, flexibility, physical validity, and computational efficiency.

\section{Control Problems}\label{sec:IC}
\begin{table*}[ht]
\centering
\scriptsize
\caption{Overview of AI methods for control problems.}
\label{tab:control}
\vspace{-0.2cm}

\setlength{\tabcolsep}{4.5mm}{
\begin{tabular}{l|c|c|c}
\hline
\textbf{Methods} & \textbf{Source} & \textbf{Backbone} & \textbf{Tasks} \\ \hline

\multicolumn{4}{c}{\textbf{\ref{IC:Opt} Optimization-based control}} \\ \hline
\multicolumn{4}{l}{\textbf{\ref{IC:Opt:PDECons} Physics-in-the-loop control}} \\ \hline
CFE~\cite{holl2020learning} & \textit{ICLR2020} & CNN & Fluid flow control \\ \hline 
Control PINNs~\cite{barry2025physics} & \textit{CAMC2025} & MLP &  Heat control \\ \hline
PhysNet~\cite{gokhale2022physics} & \textit{Applied Energy2022} & MLP & Building thermal control \\ \hline
PhysCon~\cite{chen2023physics} & \textit{Build. Environ.2023} & MLP & Building thermal control \\ \hline
Mowlavi \textit{et al.}~\cite{mowlavi2023optimal} & \textit{JCP2023} & MLP & Fluid optimal control \\ \hline
Arnold \textit{et al.}~\cite{arnold2021state} & \textit{Eng. Appl. Artif. Intell. 2021} & MLP & Burgers equation optimal control \\ \hline

\multicolumn{4}{l}{\textbf{\ref{IC:Opt:Surrogate} Learned-model-assisted control}} \\ \hline
DMDc~\cite{proctor2016dynamic} & \textit{SIADS2016} & DMD with control & Flow control \\ \hline
Korda \textit{et al.}~\cite{korda2018linear} & \textit{Automatica2018} & Koopman & Nonlinear control \\ \hline
K-ROMs~\cite{peitz2019koopman} & \textit{Automatica2019} & Koopman & Burgers / NS control \\ \hline
RK-MPC~\cite{mamakoukas2022robust} & \textit{ACC2022} & Koopman & Robust control \\ \hline
KqLMPC~\cite{calderon2021koopman} & \textit{ECC2021} & Koopman& Adaptive control \\ \hline
Morton \textit{et al.}~\cite{morton2018deep} & \textit{NeurIPS2018} & MLP & Flows control \\ \hline
Bhan \textit{et al.}~\cite{bhan2023operator} & \textit{L4DC2023} & MLP & Adaptive PDE control  \\ \hline
MS-DeepONet~\cite{de2025deep} & \textit{OJCSYS2025} & MLP & Nonlinear predictive control \\ \hline

\multicolumn{4}{c}{\textbf{\ref{IC:Amor} Offline-Trained Control}} \\ \hline
\multicolumn{4}{l}{\textbf{\ref{IC:Amor:Operator} Operator learning}} \\ \hline
OCPs~\cite{feng2025optimal} & \textit{AAAI2025} & MLP & ODE / OCP control \\ \hline
Wang \textit{et al.}~\cite{wang2021fast} & \textit{Arxiv2021} & MLP & Poisson/Heat control \\ \hline
Bhan \textit{et al.}~\cite{bhan2023neural} & \textit{IEEE Trans. Autom. Control} & MLP & PDE boundary control \\ \hline
Krstic \textit{et al.}~\cite{krstic2024neural} & \textit{Automatica2024} & MLP & Reaction–diffusion control \\ \hline
Wang \textit{et al.}~\cite{wang2025backstepping} & \textit{ACC2025} & MLP & Hyperbolic control \\ \hline
Qi \textit{et al.}~\cite{qi2024neural} & \textit{SCL2024} & MLP & Neural PDE feedback control \\ \hline
Zhang \textit{et al.}~\cite{zhang2026operator} & \textit{Automatica2026} & MLP & Hyperbolic control \\ \hline
Lamarque \textit{et al.}~\cite{lamarque2025gain} & \textit{IEEE TAC 2025} & MLP & Transport PDE control \\ \hline

\multicolumn{4}{l}{\textbf{\ref{IC:Amor:Imitation} Imitation learning}} \\ \hline
Neural-HJB~\cite{verma2025neural} & \textit{Found. Data Sci.2025} & MLP & Advection-diffusion control \\ \hline
BEAR~\cite{mulayim2025impact} & \textit{ACM e-Energy2025} & MLP & HVAC control \\ \hline

\multicolumn{4}{l}{\textbf{\ref{IC:Amor:Generative} Generative policy learning}} \\ \hline
DiffPhyCon~\cite{wei2024diffphycon} & \textit{NeurIPS2024} & Diffusion & Complex fluid Control \\ \hline
CL-diffphycon~\cite{wei2024cl} & \textit{ICLR2025} & Diffusion& Complex fluid Control\\ \hline
SafeDiffCon~\cite{hu2025uncertain} & \textit{ICML2025} & Diffusion & Nuclear fusion control \\ \hline

\multicolumn{4}{c}{\textbf{\ref{IC:RL} Reinforcement learning}} \\ \hline
MARL-DRL~\cite{guastoni2023deep} & \textit{EPJE2023} & MLP & Turbulent control \\ \hline
Rabault \textit{et al.}~\cite{rabault2019artificial} & \textit{JFM2019} & MLP & Active flow control \\ \hline
Degrave \textit{et al.}~\cite{degrave2022magnetic} & \textit{Nature2022} & MLP & Tokamak plasmas control \\ \hline
Wang \textit{et al.}~\cite{wang2023deep} & \textit{JFM2023} & MLP & Flow control \\ \hline
SmartSOD2D~\cite{font2025deep} & \textit{Nat. Commun.2025} & MLP & Turbulent separation-bubble control \\ \hline
Invariant MARL~\cite{vignon2023effective} & \textit{Phys. Fluids2023} & MLP & RBC control \\ \hline
Suarez \textit{et al.}~\cite{suarez2025flow} & \textit{Comms. Eng.2025} & MLP & Distributed flow control \\ \hline
Novati \textit{et al.}~\cite{novati2021automating} & \textit{Nat. Mach. Intell.2021} & MLP & Turbulence control \\ \hline
Kim \textit{et al.}~\cite{kim2022safe} & \textit{AIChE 2022} & MLP & Nonlinear optimal control with constraints \\ \hline

\end{tabular}}
\vspace{-0.3cm}
\end{table*}

In contrast to inverse problems and inverse design, which focus on recovering or synthesizing static variables, control problems seek to determine a time-dependent control input that steers the evolution of a PDE-governed system toward a desired objective. Such problems arise in a wide range of applications, including fluid flow control~\cite{cattafesta2011actuators}, thermal regulation~\cite{morel2001neurobat}, and nuclear reactor control~\cite{ku1992improved}. In PDE-governed settings, control problems can generally be formulated as:
\begin{equation}
\begin{aligned}
    &a^* = \arg\min_{a} \,\mathcal{J}(u,a), \\
    \text{s.t.} \quad 
    &\mathcal{F}\!\left(x,t,u,\nabla u,\nabla^2 u,\partial_t u,\ldots;\lambda_p,a\right)=0,
\end{aligned}
\end{equation}

where $a$ denotes the control input and $\mathcal{J}(u,a)$ measures how well the controlled trajectory satisfies the desired behavior. Depending on the application, $a$ may take the form of external forcing, boundary actuation, coefficient modulation, or a feedback law. In practical implementations, especially in digital control settings, $a$ is often discretized over time as a control sequence $\{a_k\}_{k=1}^{T}$.



Control problems pose several distinctive challenges beyond static inverse problem inference or design synthesis:
(1)~\textbf{High-dimensional temporal decision space}: unlike inverse problems or inverse design, where the unknown is typically a static parameter field or design variable, control problems must optimize a time-dependent input over the entire horizon. As a result, the decision space grows with the time horizon and must respect strong temporal dependencies, making naive search or sampling strategies increasingly difficult.
(2)~\textbf{Expensive PDE-constrained evaluation}: evaluating a candidate control requires solving the forward PDE over time to obtain the resulting trajectory. Consequently, both objective evaluation and gradient-based control updates can be computationally expensive, making large-scale exploration and optimization prohibitively costly.

(3)~\textbf{Additional operational constraints}: beyond objective optimization, practical control problems often involve requirements such as partial observability, real-time responsiveness, and closed-loop stability. These constraints make PDE control problems substantially more challenging than static inverse inference or design synthesis.

These challenges have motivated various learning-based strategies for PDE control problems. Accordingly, we organize the existing literature into three categories: \textit{optimization-based control}, which explicitly solves for control actions under PDE constraints, either using the full PDE model or a learned surrogate; 
\textit{offline-trained control}, which learns a direct mapping from system states to control actions without online optimization; and \textit{reinforcement learning}, which learns control policies through reward-driven interaction with the environment. Finally, we provide a dedicated discussion on \textit{temporal and feedback structure}, a supporting dimension of PDE control problems.

\subsection{Optimization-Based Control}
\label{IC:Opt}
Optimization-based control directly optimizes the control input under dynamical constraints. Existing methods can be broadly divided into two classes: \textit{physics-in-the-loop methods}, which keep the full governing PDE in the optimization loop; and \textit{learned-model-assisted methods}, which accelerate control optimization by replacing repeated PDE solvers with learned surrogate models.

\subsubsection{Physics-in-the-Loop Control}
\label{IC:Opt:PDECons}
\begin{wrapfigure}{r}{0.45\linewidth}  
    \centering
    \includegraphics[width=\linewidth]{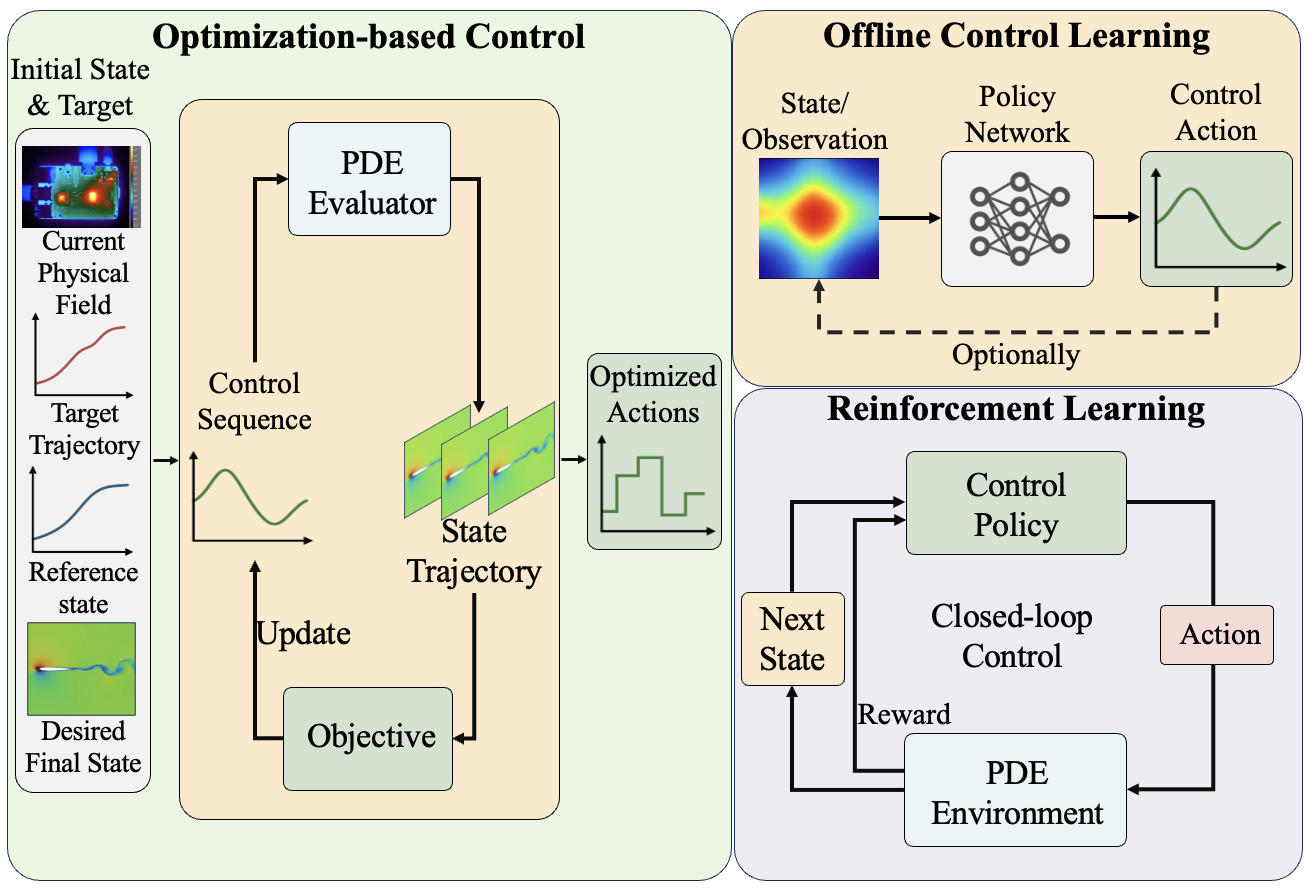}
    \caption{Overview of general AI methods for control problems.}
    \label{fig:ICoverview}
\end{wrapfigure}

Physics-in-the-loop control directly optimizes the control input under the full governing PDE. In classical settings, this is typically realized through direct-adjoint optimal control, while recent learning-based methods reformulate the same idea through physics-informed optimization or differentiable solvers, so that the system state and the control can be optimized in a more unified framework. Representative examples include~\cite{mowlavi2023optimal, arnold2021state}, which use PINNs for PDE-constrained optimal control, and~\cite{barry2025physics}, which introduces a one-stage Control-PINN framework that jointly solves for the state, adjoint, and control. Related differentiable-physics approaches, such as~\cite{holl2020learning}, further show that differentiable PDE solvers can support long-horizon control of PDE-governed systems. Overall, these methods preserve strong physical consistency, but they remain computationally demanding because the control optimization still requires repeated PDE-constrained evaluations over the time horizon.

\subsubsection{Learned-Model-Assisted Control}
\label{IC:Opt:Surrogate}
The central bottleneck of physics-in-the-loop control lies in the repeated forward solvers required at every iteration of the optimization loop. Learned-model-assisted control reduces the cost of optimization by replacing repeated PDE solvers with cheaper learned dynamics models, while preserving the overall trajectory-optimization structure. The control input is still obtained by solving an optimization problem online, but the underlying system evolution is approximated by a data-driven surrogate rather than the original PDE solver.

An influential line of work in this direction is based on data-driven reduced models such as dynamic mode decomposition (DMD) with control and Koopman learning. Early work on DMD with control~\cite{proctor2016dynamic} showed that data-driven input-output models can be extracted directly from controlled high-dimensional systems. Building on this idea,~\cite{korda2018linear} introduced Koopman-based linear predictors for model predictive control, demonstrating that nonlinear control can be made much more tractable in a lifted space. This framework was later extended to broader PDE systems in~\cite{peitz2019koopman}, and further developed toward more robust and adaptive settings in~\cite{mamakoukas2022robust,calderon2021koopman}.

Beyond Koopman-based approaches, more flexible neural surrogates have also been introduced into the control loop. For example,~\cite{morton2018deep} learns recurrent dynamical models for unsteady flow control, while recent operator-learning-based approaches further replace the predictive model with neural operators for faster control-oriented computation~\cite{bhan2023operator,de2025deep}. These methods can substantially improve computational efficiency, but their performance depends critically on whether the learned predictor remains accurate and stable over the control horizon, especially under distribution shifts.

\subsection{Offline-Trained Control}
\label{IC:Amor}
Offline-trained control shifts the computational burden from online optimization to offline training. Rather than solving a new control problem for each instance, it learns from a family of PDE control tasks and directly outputs control actions or feedback laws for deployment. We organize existing methods into three categories: \textit{operator learning}, \textit{imitation learning}, and \textit{generative policy learning}.

\subsubsection{Operator Learning}
\label{IC:Amor:Operator}
Operator-learning-based control learns a direct map from PDE states, coefficients, or boundary conditions to control trajectories or feedback gains, thereby enabling fast deployment without online optimization. One line of work learns the solution map for PDE-constrained optimal control itself, enabling the direct prediction of optimal trajectories or control laws for new instances. Foundational neural-operator architectures such as~\cite{lu2019deeponet,li2020fourier} provide the basis for this direction, while more control-specific works, including~\cite{feng2025optimal,wang2021fast}, further show that mappings from PDE parameters to optimal controls can be learned offline and reused efficiently.

A second line applies operator learning to approximate stabilizing feedback laws derived from classical PDE backstepping~\cite{zhou2008adaptive}. In this setting, operator learning replaces repeated gain or kernel computation by directly learning the map from PDE coefficients or boundary conditions to backstepping controllers. Representative works include~\cite{bhan2023neural,krstic2024neural,wang2025backstepping,qi2024neural}, with related extensions considering robust stabilization, adaptive control, and gain scheduling~\cite{zhang2026operator,bhan2023operator,lamarque2025gain}. The main advantage of operator learning is its extremely low deployment cost once training is complete, although its performance may degrade when the test-time PDE parameters or operating conditions move beyond the training distribution.

\subsubsection{Imitation Learning}
\label{IC:Amor:Imitation}
Imitation learning learns a state-to-action policy from offline demonstrations by imitating an expensive reference controller~\cite{bain1995framework,hussein2017imitation}. Unlike operator learning, which predicts full trajectories or feedback gains from PDE-level descriptions, imitation learning focuses on directly reproducing expert control behavior from state-action pairs. The reference controller may come from PDE-constrained optimization, or collected high-fidelity data, and the policy is trained to reproduce its behavior through supervised regression. Once trained, deployment reduces to a single forward pass, with a cost independent of the underlying PDE solver. A representative example is~\cite{verma2025neural}, which learns parameterized control laws in a supervised setting and demonstrates fast deployment given sufficiently rich offline data. Related hybrid efforts have also explored combining imitation-based initialization with subsequent online adaptation. For example,~\cite{mulayim2025impact} studies HVAC control through a physics-informed reinforcement learning framework that starts from imitation and then performs online fine-tuning. Overall, imitation learning offers a simple and efficient route to fast control inference, although its performance depends strongly on the quality and coverage of the expert data.

\subsubsection{Generative Policy Learning}
\label{IC:Amor:Generative}
Beyond deterministic imitation, recent work has extended offline control learning to generative modeling of control signals, where the policy is represented as a conditional distribution over control trajectories rather than a single regression output. In this setting, control is cast as a conditional generation problem, enabling the model to represent multimodality in the space of feasible actions. Representative examples include~\cite{wei2024diffphycon}, which formulates control using diffusion-based conditional generation, and~\cite{wei2024cl}, which incorporates feedback into the diffusion process to enable closed-loop control. In addition,~\cite{hu2025uncertain} integrates uncertainty quantification into diffusion-based control generation, thereby improving robustness. These methods retain the low deployment cost of offline learning while providing a more flexible representation of feasible control behaviors.

\subsection{Reinforcement Learning}
\label{IC:RL}
Reinforcement learning learns the control policy directly through interaction with the environment, rather than relying on an explicit model or an offline optimization procedure~\cite{kaelbling1996reinforcement, arulkumaran2017deep, schulman2017proximal, konda1999actor}. In PDE-governed settings, the system state or sensor observation serves as the policy input, the action corresponds to the control signal, and the reward is defined by the control objective. Early representative demonstrations in PDE settings include~\cite{rabault2019artificial}, which showed that reinforcement learning can discover control strategies for cylinder flow, and~\cite{degrave2022magnetic}, which demonstrated direct deployment of a learned policy on a real fusion device. Since then, reinforcement learning has been extended to increasingly challenging regimes, including turbulent drag reduction in channel flows~\cite{guastoni2023deep}, active flow control in turbulent separation bubbles~\cite{font2025deep}, and transfer of learned policies to high Reynolds number settings~\cite{wang2023deep}. Across these works, reinforcement learning proves particularly effective in strongly nonlinear, turbulent, or partially observed settings, where repeated online optimization becomes impractical.

Recent developments have focused on improving scalability, sample efficiency, and safety. To handle large action spaces and spatially distributed actuation, several works adopt multi-agent formulations, where each actuator or subdomain is treated as an independent agent, enabling decentralized control of complex flow configurations~\cite{vignon2023effective, suarez2025flow, novati2021automating}. In parallel, physics-informed and safe variants have been proposed to improve reliability and constraint satisfaction: physics-informed approaches embed PDE structure into the reward or policy update to improve sample efficiency~\cite{wang2024physics}, while safe reinforcement learning methods incorporate explicit state and input constraints to guarantee feasibility during both training and deployment~\cite{kim2022safe}.

\subsection{Temporal and Feedback Structure of PDE Control}
Beyond the optimization pipeline, PDE control problems also depend critically on their temporal structure. Unlike inverse problems and inverse design, control must determine actions over time rather than solve for a single static variable. From this perspective, existing methods mainly differ in how they handle decision making over the time horizon.

A first class follows trajectory-level planning, where the objective is to optimize or generate a full control sequence over a finite horizon. This formulation is common in PDE-constrained optimal control, and also appears in recent generative control methods that model control trajectories over the whole horizon~\cite{wei2024diffphycon}. A second class adopts receding-horizon control, where a predictive model is used to optimize a short-horizon sequence, but only the first action is executed before re-optimization at the next time step. This setting is typical in model predictive control and learned surrogate control, including Koopman-based approaches~\cite{korda2018linear,peitz2019koopman}. A third class learns feedback policies, which directly map the current state or observation to the next control action. This setting is common in imitation learning and reinforcement learning, and can also arise in operator-learning-based approximations of feedback laws~\cite{rabault2019artificial,verma2025neural,qi2024neural, wei2024cl}.




\section{Applications}\label{sec:APP}

\subsection{Mechanical Systems}
\label{APP:Mechanical}
Mechanical systems form an important class of inverse PDE problems, where the goal is to recover material parameters, constitutive properties, or structural configurations from observable mechanical responses such as displacement, strain, or load fields. Recent learning-based methods mainly advance this area along three directions. First, physics-informed approaches use PDE constraints for parameter identification and field reconstruction, as demonstrated in solid mechanics\cite{haghighat2021physics}. Second, neural networks have been used to accelerate topology optimization, for example by predicting sensitivity information or embedding online learning into the optimization loop \cite{senhora2022machine,deng2022self}. Third, generative models have become increasingly important for inverse design of mechanical metamaterials, where diffusion-based methods can generate diverse microstructures conditioned on target stress-strain responses ~\cite{yang2024guided,bastek2023inverse}. Together, these works show that inverse problems in mechanical systems are evolving from parameter recovery toward accelerated topology optimization and generative structural design.

\subsection{Aerodynamics}
\label{APP:Aerodynamic}
Aerodynamic inverse design aims to find geometries that achieve desired aerodynamic responses, such as prescribed pressure distributions or reduced drag. Since repeated CFD evaluations are costly, recent learning-based methods have developed more efficient design pipelines, including direct inverse mappings from target responses to airfoil geometries~\cite{balla2022inverse,anand2024novel}, surrogate-assisted optimization with differentiable CFD solvers and neural operators~\cite{bezgin2023jax,shukla2024deep}, and large-scale data-driven benchmarks such as the DrivAerNet series~\cite{elrefaie2024drivaernet,elrefaie2024drivaernet++}. To address ambiguity and multimodality in the design space, recent work further explores invertible neural networks and reinforcement learning for aerodynamic shape optimization~\cite{glaws2022invertible,dussauge2023reinforcement}. Related studies also extend inverse design beyond static geometry optimization to active flow control, where learned policies regulate aerodynamic behavior across airfoils~\cite{zheng2024transformer,zha2007high}.

\subsection{Thermal Systems}
\label{APP:Thermal}
Thermal systems form a broad class of inverse PDE problems in building energy systems, thermal management, and industrial heat transfer, where the goal is to infer hidden quantities such as heat sources, thermal diffusivity, boundary heat fluxes, or control-oriented state models from sparse temperature observations governed by heat conduction or diffusion equations. These problems are typically ill-posed, noise-sensitive, and computationally expensive to solve repeatedly. Recent learning-based methods, especially physics-informed neural networks, have been used to recover thermal diffusivity, inaccessible boundary fluxes, and thermophysical parameters from sparse measurements, with validation against finite element and analytical solutions~\cite{cai2021physics,billah2023physics,xu2023physics}. In building applications, physics-informed models improve data efficiency and long-horizon prediction for control-oriented thermal modeling~\cite{gokhale2022physics}. Related applications further extend to electronic thermal management~\cite{zhang2024physics} and industrial heat-transfer monitoring and regulation~\cite{majumdar2025hxpinn}.

\subsection{Full-waveform Inversion for Geophysical Imaging}
\label{APP:Full-waveform}
Full-waveform inversion (FWI) is a representative inverse PDE problem in geophysics that seeks to recover subsurface parameters, such as velocity, from observed seismic waveforms by matching them with synthetic data generated by wave equations \cite{zhang2020data,tang2021deep}. While FWI offers high-resolution imaging, it remains highly nonlinear and ill-posed, with major challenges including cycle-skipping, noise sensitivity, and strong dependence on the initial model. Recent learning-based methods have substantially advanced FWI. Inverse-mapping approaches such as InversionNet learn the inverse map from seismic data to velocity models, significantly reducing iterative cost \cite{wu2019inversionnet}. Large-scale studies further show the importance of dataset diversity and benchmarks such as OpenFWI for improving generalization \cite{li2025towards}. More recent works incorporate stronger architectural and prior designs, including vision Transformers, implicit neural representations, and noise-aware training, to improve inversion under complex structures and noisy conditions \cite{kang2026implicit,zhang2025lateral,zhang2026enhancing}.

\subsection{System Identification in Dynamical Systems}
\label{APP:Dynamical}


System identification in dynamical systems can be regarded as an inverse PDE problem that aims to recover unknown parameters or latent dynamics from partial temporal observations. Recent learning-based methods move beyond static inverse mappings by combining differentiable simulation with parameter estimation, enabling hidden physical parameters to be identified through backpropagation in real-time control~\cite{chen2022real} and robotic manipulation of elastoplastic materials from sparse trajectories and incomplete point clouds~\cite{yang2025differentiable}. Recent studies further extend this paradigm from parameter recovery to governing-equation discovery and high-dimensional observation modeling: latent-state reconstruction with sparse regression can uncover interpretable dynamics in partially observed systems~\cite{lu2022discovering}, while PAC-NeRF couples neural radiance fields with differentiable continuum mechanics to jointly infer geometry and physical parameters from multi-view videos~\cite{li2023pac}. These works indicate a shift toward unified inverse PDE system identification frameworks that integrate differentiable physics, equation discovery, and neural field based observations.

\subsection{Medical Imaging Reconstruction}
\label{APP:Medical}
Medical imaging reconstruction is a representative inverse PDE problem that seeks to recover images or physical fields from incomplete, noisy, or indirect measurements governed by acquisition physics, such as the Radon transform in CT and Fourier encoding in MRI. Recent methods increasingly use diffusion and score-based generative priors to regularize such ill-posed reconstructions. Early work showed that score models trained only on medical images can be combined with measurement operators for unsupervised CT and MRI reconstruction~\cite{songsolving}, while $\Pi$GDM introduced pseudoinverse-guided conditional scores for task-agnostic diffusion-based inversion~\cite{song2023pseudoinverse}. Subsequent methods extended this paradigm to practical 3D reconstruction, including sparse-view and limited-angle CT, compressed-sensing MRI~\cite{chung2023solving}, limited-angle CT with model-based conditional diffusion~\cite{liu2023dolce}, latent diffusion with hard data consistency~\cite{song2023solving}, and scalable multi-coil MRI and 3D CT via Krylov-subspace updates~\cite{chung2023decomposed}. Overall, these works shift medical inverse reconstruction from supervised image-to-image regression toward posterior sampling frameworks that couple learned generative priors with physics-based data consistency.


\section{Challenges and Future Prospects}\label{sec:Challenges}

\subsection{Physics Informed Architectures}
A central challenge in physics-informed learning is that most existing approaches inject physical laws through auxiliary loss terms, such as PDE residual, boundary-condition, or conservation penalties, rather than through the model architecture itself \cite{raissi2020hidden,finzi2023stable}. This soft-constraint paradigm is appealing for its generality, but it offers no guarantee that the imposed physics residual can be optimized to zero. More importantly, a small residual on sampled collocation points does not necessarily imply faithful satisfaction of the governing equations over the full domain. This issue becomes more severe in time-dependent problems, where global residual minimization may conflict with the causal structure of evolution equations and suffer from optimization pathologies such as catastrophic forgetting, whereas local-in-time alternatives can introduce ill-conditioning and unfavorable scaling with model size \cite{finzi2023stable}. These limitations suggest that physics injection through losses alone may be insufficient for robust and scalable scientific learning.

An emerging alternative is to encode physical structure directly into the network architecture, for example by embedding discretization rules, conservation mechanisms, or PDE constituents into the network design \cite{karlbauer2022composing,rao2023encoding}. Such models reduce the burden on optimization to ``discover'' physics from penalties alone and often exhibit improved generalization, interpretability, and data efficiency. However, architecture-level physics injection remains difficult in realistic settings, where the governing equations may be incomplete, closure relations may be unknown, and the model must still adapt to varying geometries, boundary conditions, and multiscale couplings. If the inductive bias is too rigid, the architecture may enforce an incorrect physical structure and degrade flexibility. Future work should therefore focus on hybrid designs that hard-code only trusted principles while leaving uncertain components learnable, as well as on developing architectures with built-in stability, conservation, and operator-level consistency, supported by stronger theoretical guarantees on optimization, identifiability, and out-of-distribution generalization.

\subsection{Uncertainty Quantification}
Inverse PDE problems aim to recover hidden quantities from incomplete, noisy, and indirect observations, making the recovered solution inherently non-unique and statistically ambiguous. Such uncertainty arises from measurement noise, sparse or indirect observations, ill-posed inverse mappings, mismatch between the assumed PDE model and the true physical system, as well as surrogate-model errors and distribution shifts in learning-based solvers. In such settings, a deterministic point estimate may be insufficient, because multiple plausible solutions can explain the same observation. Recent studies have explored uncertainty modeling from several directions. Bayesian methods provide a principled framework for posterior uncertainty in PDE-governed inverse problems~\cite{bui2012extreme}. Bayesian neural operators extend Bayesian inference to operator learning and provide posterior estimates over PDE solution operators in function spaces~\cite{magnaniapproximate}. Related uncertainty-aware PINN and neural-operator methods address uncertainty under noisy input-output settings~\cite{zou2025uncertainty}. Ensemble-based methods estimate uncertainty by aggregating predictions from multiple models~\cite{mouli2024using,wu2024uncertainty}. Diffusion and score-based models provide a flexible framework for posterior sampling. For example, DiffusionPDE~\cite{huang2024diffusionpde} learns the joint distribution of coefficient and solution fields, while ODE-based diffusion posterior sampling~\cite{jiang2024ode} incorporates score-based priors into PDE-constrained inverse inference. In addition, conformal prediction for operator learning~\cite{ma2024calibrated} and physics-informed conformal prediction~\cite{gopakumar2025calibrated} provide calibration-oriented uncertainty estimates with distribution-free coverage guarantees. These methods avoid collapsing the inverse solution into a single prediction and can provide multiple candidate reconstructions, uncertainty intervals, or calibrated prediction sets consistent with observations and physical constraints.

However, uncertainty quantification in AI-based methods for inverse PDE problems remains far from mature. Existing methods often address only part of the uncertainty sources, such as posterior variability, predictive disagreement, or calibration error, while realistic inverse pipelines involve their interactions across observation, modeling, surrogate prediction, and downstream optimization. Future work should therefore develop physics-aware and task-aware uncertainty quantification methods that jointly account for measurement noise, inverse non-uniqueness, surrogate-model error, model-form mismatch, and uncertain physical constraints. Ultimately, reliable inverse PDE solvers should provide calibrated and interpretable confidence estimates, quantifying not only plausible solutions but also their trustworthiness under the assumed physical model and downstream task requirements.

\subsection{Inverse Foundation Model}
Although foundation models have achieved remarkable success in natural language processing and computer vision, most AI-based inverse PDE methods remain tailored to specific equations, domains, geometries, or objectives. As a result, a model trained for one task, often fails to transfer to other tasks. Similarly, models trained on fixed grids or sensing patterns often struggle to generalize across resolutions, boundary conditions, geometries, and observation layouts. This task-specific nature limits the scalability of current inverse PDE methods and makes it difficult to build reusable AI systems for scientific discovery and engineering design. Recent progress on forward PDE foundation models provides useful inspiration. UniSolver~\cite{zhou2024unisolver} incorporates equation symbols, coefficients, and boundary conditions as conditioning information for cross-PDE generalization. PROSE-FD~\cite{liu2024prose} jointly learns forward state prediction and symbolic equation discovery, while DPOT~\cite{hao2024dpot} shows that denoising-based pretraining can improve operator-level generalization across multiple PDEs. However, these models mainly focus on forward prediction, and foundation models specifically designed for inverse problems remain largely absent.

The goal of inverse foundation models is to move from isolated task-specific solvers toward general-purpose inverse solving systems. Rather than learning a separate inverse map for each problem, such models should support diverse conditioning signals, including observations, target responses, design constraints, and control objectives, so that a single model can adapt to multiple inverse tasks. Recent efforts such as FluidZero~\cite{fengfluidzero} provide an early example by using a unified generative model for fluid-structure interaction tasks, including prediction, parameter identification, design, and control problems. Building such models requires architectures capable of handling heterogeneous scientific representations, such as grids, meshes, graphs, point clouds, and neural fields, together with unified training objectives that capture the shared goal underlying diverse inverse problems. 

\subsection{Limited Real-world Data}

In the era of foundation models, scaling laws suggest that increasing model size and data scale can consistently improve performance and generalization. For PDE systems, several benchmark datasets have been introduced, including PDEBench~\cite{takamoto2022pdebench} and PINNacle~\cite{hao2024pinnacle} for time-dependent PDE tasks; CFDBench~\cite{luo2023cfdbench} and NASA TMR~\cite{rumsey2017nasa} for computational fluid dynamics and turbulence modeling; and larger-scale benchmarks such as FlowBench~\cite{tali2024flowbench}, The Well~\cite{ohana2024well}, and BLASTNet~\cite{chung2022blastnet} for flow simulation, spatiotemporal physical systems, and reacting flows. Readers may refer to~\cite{wu2025physics} for a more comprehensive overview. However, unlike the visual domain, which benefits from abundant large-scale real-world data, most existing PDE benchmarks are generated by numerical simulators, creating a nontrivial gap between training data and real-world scenarios. This gap limits the applicability of large models in scientific settings and often leads to poor generalization.

One promising way to mitigate this limitation is to construct paired real-to-simulation datasets. For example, RealPDEBench~\cite{hu2026realpdebench} couples real-world observations with simulated data to narrow the gap between simulation and reality. Another direction is to exploit the benefits of pretraining. GeoPT~\cite{wu2026geopt} first pretrains a model on a large-scale geometric dataset with simulated velocity fields, and then fine-tunes it on a specific target domain with limited real-world data, leading to improved performance. Ultimately, however, data-hungry AI methods for scientific problems will require large-scale real-world datasets. Achieving this goal will depend on closer collaboration across scientific disciplines and stronger connections between AI and domain-specific communities.

\section{Conclusion}
This survey provides the first systematic review of recent advances in applying artificial intelligence to inverse PDE problems. We have presented a unified perspective that organizes the rapidly growing literature into three major categories: inverse problems, inverse design, and control problems, and have summarized representative learning-based paradigms developed for each setting. A broad range of scientific and industrial applications has also been reviewed. More importantly, several open challenges have been highlighted, including physics-informed architectural design, uncertainty quantification, inverse foundation models, and the limited availability of real-world datasets. We further outline several promising future directions for addressing these challenges. Overall, the progress reviewed in this paper suggests that AI is not merely an auxiliary tool for inverse PDE problems but is increasingly becoming a vital paradigm for improving efficiency, scalability, and adaptability in solving complex inverse problems.

\bibliographystyle{ACM-Reference-Format}
\bibliography{sample-base}

\end{document}